%% file: main.tex
\documentclass[conference]{IEEEtran}
\usepackage{times}

\usepackage[numbers]{natbib}
\usepackage{multicol}
\usepackage[bookmarks=true]{hyperref}
\usepackage[binary-units=true]{siunitx}
\usepackage{times}
\usepackage{amsmath,amssymb}

\usepackage{accents}
\usepackage{graphicx}
\usepackage{bm}
\let\labelindent\relax
\usepackage[inline]{enumitem}
\setlist[itemize,1]{leftmargin=0.45cm}
\usepackage{array}
\usepackage{multicol}
\usepackage{multirow}
\usepackage{tabulary}
\usepackage[table, dvipsnames]{xcolor}
\usepackage{booktabs}
\usepackage{todonotes}
\usepackage{cite}
\usepackage{algorithm}% http://ctan.org/pkg/algorithm
\usepackage{algpseudocode}% http://ctan.org/pkg/algorithmicx
\usepackage{breqn}
\usepackage{tcolorbox}
\usepackage{mathtools}
\usepackage{cuted}
\usepackage{stfloats}
\usepackage{balance}
\usepackage{caption}
\usepackage{graphicx}
\usepackage{subcaption}

\usepackage{tcolorbox}
\usepackage{mathtools}
\usepackage{tikz}
\usetikzlibrary{arrows}
\usetikzlibrary{calc}
\usetikzlibrary{arrows,decorations.markings}
\usepackage{tkz-euclide}
\usepackage{nomencl}
\usepackage{pgfplots}
\usetikzlibrary{arrows.meta}
\usepackage{nicefrac}

\usepackage{overpic}
\usepackage{nomencl}
\usepackage{xpatch}

\xpatchcmd{\thenomenclature}{%
	\section*{\nomname}% Look for `\section*... etc.
}{% Replace it by 'nothing'
}{\typeout{Success}}{\typeout{Failure}}
\usetikzlibrary{patterns}

\newcommand{\stripedbox}[1]{%
	\tikz[baseline=(X.base)]{
		\node[
		inner sep=1pt,
		pattern=crosshatch,
		pattern color=gray!20,
		draw=gray!30
		] (X) {#1};
	}
}

\makenomenclature

\renewcommand{\thetable}{\arabic{table}}
\allowdisplaybreaks

\input{./commands.tex}

\makeatletter
\newcommand{\thickhline}{%
	\noalign {\ifnum 0=`}\fi \hrule height 1pt
	\futurelet \reserved@a \@xhline
}
\newcolumntype{"}{@{\hskip\tabcolsep\vrule width 1pt\hskip\tabcolsep}}
\makeatother

\begin{document}

% paper title
\title{\LARGE \bf Integrated Hierarchical Decision-Making in Inverse Kinematic Planning and Control}
\author{
	\authorblockN{Kai Pfeiffer\authorrefmark{1},
		Quan Zhang\authorrefmark{1},
		Yuqing Chen\authorrefmark{1},
		Gordon Owusu Boateng\authorrefmark{1},
		Yuquan Wang\authorrefmark{2},
		Vincent Bonnet\authorrefmark{3}\authorrefmark{4},\\
		Abderrahmane Kheddar\authorrefmark{5}		
	}
	\authorblockA{\authorrefmark{1}School of Advanced Technology,
		Xi'an Jiatong Liverpool University,
		China \quad \authorrefmark{2}X Square Robot, China}
		\authorblockA{
			\authorrefmark{3} CNRS, IPAL, Singapore \quad\authorrefmark{4}LAAS-CNRS, Universit\'e de Toulouse, France}
			\authorblockA{ \authorrefmark{5}CNRS-University of Montpellier, LIRMM, Montpellier, France}
}

	\maketitle
\thispagestyle{empty}
\pagestyle{empty}

% tldr: This paper proposes an efficient nonlinear optimization method for inverse kinematic planning and control with simultaneous hierarchical decision-making in the presence of many possible candidate solutions.

\begin{abstract}%
This work presents a novel and efficient nonlinear programming framework that tightly integrates hierarchical decision-making with whole-body inverse kinematic planning and control.
Decision-making plays a central role in many aspects of robotics, from sparse inverse kinematic control with a minimal number of joints, to inverse kinematic planning while simultaneously selecting a discrete end-effector location from multiple candidates. Current approaches often rely on heavy computations using mixed-integer nonlinear programming, separate decision-making from inverse kinematics (some times approximated by reachability methods), or employ efficient but less versatile $\ell_1$-norm formulations of linear sparse programming, without addressing the underlying nonlinear problem formulations. 
In contrast, the proposed sparse hierarchical nonlinear programming solver is efficient, versatile, and accurate by exploiting sparse hierarchical structure and leveraging the $\ell_0$-norm which is rarely used in robotics. 
The solver efficiently tackles complex nonlinear hierarchical decision-making problems  previously unaddressed in the literature, such as inverse kinematic planning with simultaneous prioritized selection of end-effector locations from a large set of candidates, or inverse kinematic control with simultaneous selection of bi-manual grasp locations on a randomly rotated box. 
\end{abstract}

\section{Introduction}
Robotics often involves aspects of decision-making, for example in inverse kinematics (IK) control with a minimal number of active joints (e.g., in highly redundant robotic system embedded with brakes at the actuator level).
%Robots typically possess more degrees of freedom than are required for a given kinematic task.
Exploiting sparsity in such problems can lead to more economical~\citep{Goncalves2015} and human-like motions~\citep{Berret2008}.
% , and increases joint fault tolerance~\citep{faultTolerant2022}.
% KP: just for confirmation that you removed this reference on purpose
Furthermore, IK planning may involve multiple feasible scenarios, for instance when a robot may place its end-effector at any of several possible discrete locations without compromising the overall objective. 
While optimization methods like mixed-integer nonlinear programming (MINLP) can solve such decision-making problems to global optimality~\citep{scip_minlp,Shirai2022}, they are computationally taxing. A more efficient approach is to rely on sparse optimization methods, as proposed in~\citep{Song2021} for footstep planning with robustness to sparse gradients, unlike reinforcement learning based methods~\citep{beamdojo}. While the use of the $\ell_1$-norm in~\citep{Song2021} is effective for this purpose, some inaccuracies remain, and redundant allocation of end-effector locations was reported.
Furthermore, kinematic reachability approximations are used while the robot’s whole-body IK are treated separately. This introduces the risk that a selected end-effector location may be unreachable if the approximation is not conservative enough, or overly conservative if chosen too restrictively, thus underutilizing the robot’s workspace.
\begin{figure}[tp!]
	\centering
	\begin{overpic}[height=3.6cm]{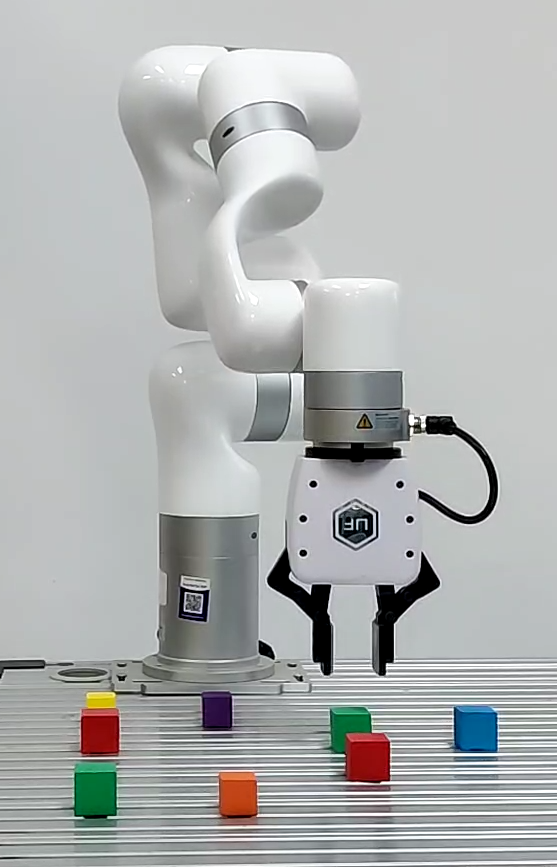}
		\put(1,89.5){\color{white}\textbf{(a)}}
	\end{overpic}\begin{overpic}[height=3.6cm]{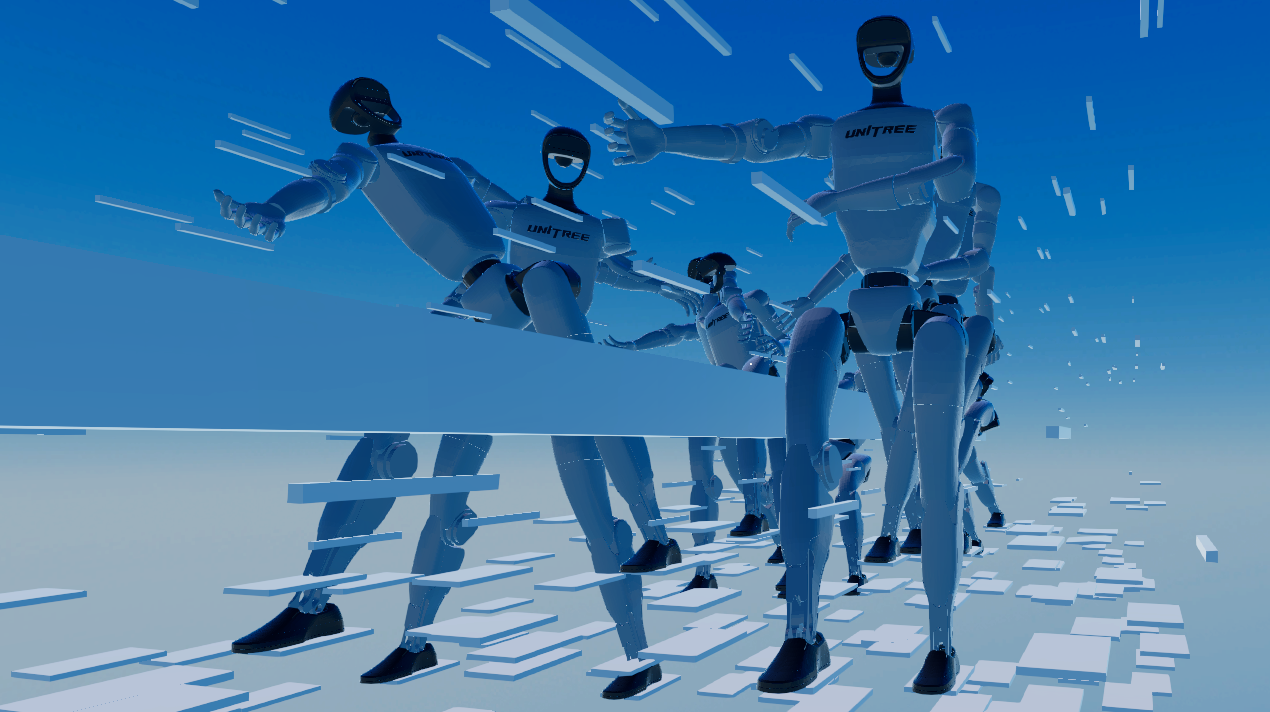}
		\put(2,50){\color{white}\textbf{(b)}}
	\end{overpic}
	
		\vspace{3pt}
		
	\begin{overpic}[height=2.98cm]{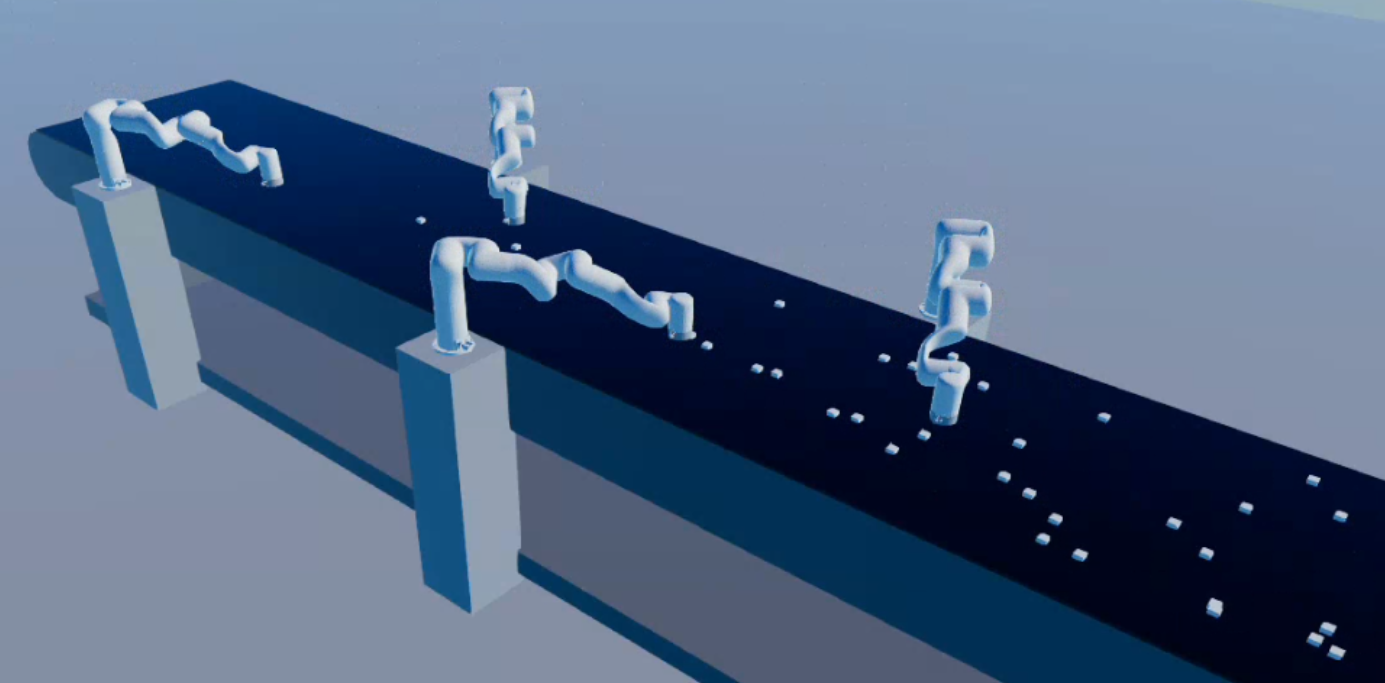}
		\put(1,3){\color{white}\textbf{(c)}}
	\end{overpic}\begin{overpic}[height=2.98cm]{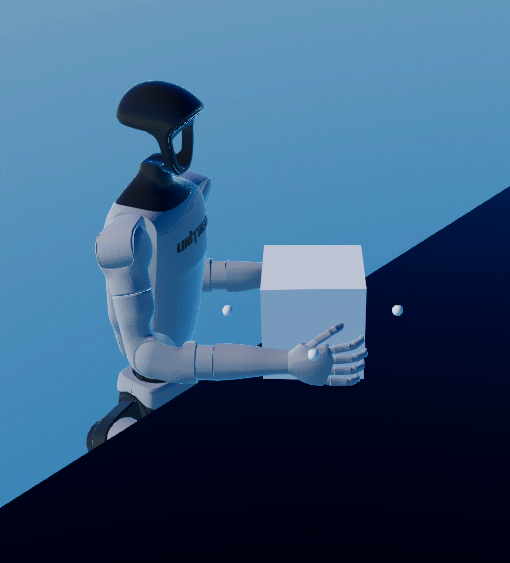}
		\put(3,6){\color{white}\textbf{(d)}}
	\end{overpic}

	\captionof{figure}{The newly proposed sparse hierarchical nonlinear programming framework (SH-NLP) enables the following use-cases: \textbf{(a)} Subsequent pick-and-place of several objects with UFactory's \texttt{xarm6}   (Sec.~\ref{sec:shikpxarm6}). \textbf{(b)} Unitree's \texttt{G1} plans IK postures while simultaneously choosing one of 200 possible discrete locations for the higher-priority foot and the lower-priority hand placement tasks. 10 postures are computed by iteratively solving 763 SHQP's within 2.12~s by the proposed solver $\mathcal{N}$\hspace{-2pt}QP (Sec.~\ref{sec:eval:plan}).  Both \textbf{(a)} and \textbf{(b)} represent planning applications (\mbox{SHIK-P}). \textbf{(c)} An array of four UFactory \texttt{xarm6}'s equipped with vacuum tubes clears a conveyor belt loaded with produce like cashew nuts. \textbf{(d)} A \texttt{G1} places its hands on two of the four sides of a randomly rotated box. In both \textbf{(c)} and \textbf{(d)}, the robot reacts to changes of the objects with immediate and continuous decision-making within the IK control loop (\mbox{SHIK-C}; see Sec.~\ref{sec:robotics}).  Rendering: \texttt{Raisim}~\citep{raisim}.}
	\label{fig:videosims}
\end{figure}

This work addresses these limitations by developing a novel framework for hierarchical nonlinear decision-making, which, to the best of our knowledge, is the first in robotics (and beyond) to address such problems. The following use-cases are considered, as illustrated in Fig.~\ref{fig:videosims}: 
\begin{itemize}
\item \emph{Planning (SHIK-P):} directly solving nonlinear sparse hierarchical IK and decision-making planning problems.
\item \emph{Control (SHIK-C):} solving sparse hierarchical IK and decision-making instantaneously for real-time control.
\end{itemize}

The main contributions of this work are as follows:
\begin{itemize}
\item We propose a novel and efficient sequential sparse hierarchical quadratic programming solver (S-SHQP) for sparse hierarchical nonlinear programs (SH-NLP). This enables robot IK planning and control with autonomous and reliable hierarchical decision-making, for example when choosing among multiple Cartesian locations with different prioritized end-effector tasks. Prioritization overcomes the difficulty of weighting tasks against logarithmic surrogate functions of the used $\ell_0$-norm formulations. 
\item Our method enables real-time decision-making in milliseconds for complex whole-body robotic systems due to linear dependency on the number of candidates. This enables efficient selection from a large set of candidate locations without relying on reachability approximations. By simultaneously considering whole-body IK, an immediate certificate of kinematic reachability is obtained.
\item Unlike sparse programming methods based on the $\ell_1$-norm, our $\ell_0$-norm based formulation is versatile and agnostic to the specific formulation of the decision-functions. For example, it enables decision-making with vector-valued functions which have been turned into scalars by least-squares. 
\item Our method enables parallel decision-making from the same set of candidates while avoiding double-allocation. This is important in robotics as end-effectors like the left and right feet typically choose from the same set of candidate locations.
\item A wide range of whole-body IK planning and control applications with integrated hierarchical decision-making for both humanoids and manipulators are presented, while reiterating on the advantages of the $\ell_0$-norm over the $\ell_1$-norm and MINLP.
\end{itemize} 

%Hierarchical optimization offers an intuitive way to structure layered objectives without weighting; it is widely used in whole-body kinematic~\citep{escande2014} and dynamic control~\citep{pfeiffer2023}.

 S-SHQP extends nonlinear hierarchical least-squares programming ($\ell_2$-norm~\citep{pfeiffer2023,pfeiffer2024}), which is solved via sequential hierarchical least-squares programming using a hierarchical step filter (HSF) and a trust-region constraint~\citep{pfeiffer2024}. Here, we introduce the required adaptations for $\ell_0$-norm programming and hierarchical decision-making (see Sec.~\ref{sec:shsqp} and~\ref{sec:hdm}).
In its main computation step, S-SHQP iteratively solves sparse hierarchical quadratic programs (SHQP). While standard QP solvers such as PIQP~\citep{piqp} or MOSEK~\citep{mosek} can be used, they do not exploit the special structure of SHQP. Our proposed solver, $\mathcal{N}$\hspace{-2pt}QP, leverages this structure for improved computational efficiency (Sec.~\ref{sec:ipm}). For example, SHIK-C with selection among $100$ possible locations on Unitree’s \texttt{G1} robot is run at a control loop time of $1.5$~ms, compared to $2.2$~ms and $8.3$~ms for PIQP and MOSEK, respectively (Sec.~\ref{sec:eval:continSelect}).

The remainder of this letter is organized as follows. Sec.~\ref{sec:relwork} reviews related work, while Sec.~\ref{sec:sparseProgDecMak} recalls the basics of sparse programming based decision-making. This is extended to hierarchical decision-making via SH-NLP, which is outlined in Sec.~\ref{sec:shnlp}. Sec.~\ref{sec:shsqp} presents the S-SHQP solver, followed in Sec.~\ref{sec:hdm} by additions for parallel and hierarchical decision-making. Sec.~\ref{sec:ipm} details the interior-point method for SHQP, and Sec.~\ref{sec:eval} evaluates the algorithm on benchmark functions and robot planning and control problems.\\

\section*{Nomenclature}
%\footnotesize
\small
\nomenclature[01]{\(\chi/x\in \mathbb{R}^n,\nu/v,\tau/t\)}{SH-NLP and SHQP primal variable equivalents}
\nomenclature[02]{\(\bullet^*\in \mathbb{R}^n\)}{Optimal SH-NLP or SHQP value $\bullet$ found for level $l$}
\nomenclature[03]{\(\bC_l\in\mathbb{R}^{\vert{\bC_l}\vert}\)}{Constraint set composed of equality and inequality constraints $\bC_l=\{\bE_l,\bI_l\}$ of level $l$}
\nomenclature[04]{\(\mI_{\cup l-1}/\mA_{\cup l-1}\)}{Set union $\mI_{\cup l-1}\coloneqq \mI_{1}\cup \dots \cup \mI_{l-1}$ of inactive / active constraints}
\nomenclature[05]{\(\mathcal{T}_{l}\in\mathbb{R}^{2\vert{\bC_l}\vert}\)}{Set of $\ell_0$-norm auxiliary constraints}
\nomenclature[06]{\(\omega_{\bC_l}^{(*)}\in\mathbb{R}^{\vert{\bC_l}\vert}\)}{$\ell_0$-weights; (*): optimal SH-NLP value found for level $l$}
\nomenclature[07]{\(f_{i/j/k}\in\mathbb{R}\)}{nonlinear functions associated with constraint sets $i\in\bC_l$ / $j\in\mI_{\cup l-1}$ / $k\in\mA_{\cup l-1}$}
\nomenclature[08]{\((f_i)\in\mathbb{R}^{\vert{\bC_l}\vert}\)}{Vector accumulation $(f_i)\coloneqq(f_i)_{i\in\bC_l}$ or $(f_i)\coloneqq[ f_{i_1} \hspace{2pt} \dots]^T$}
\nomenclature[09]{\(a_i\in\mathbb{R}^n,b_i\in\mathbb{R}\)}{Linearization of nonlinear function $f_i$}
\nomenclature[10]{\(N_{l}\in\mathbb{R}^{n\times n_r}\)}{Nullspace of matrix $[ a_{k_1}^T \hspace{2pt}\dots]^T$ of active constraints $\mathcal{A}_l$}
%\nomenclature[11]{\({z}_{l}\in \mathbb{R}^{n_r}\)}{Projected primal}
\nomenclature[12]{\(\tilde{a}_i\in\mathbb{R}^{n_r}\)}{Nullspace projected vector $\tilde{a}_i^T=a_i^TN_{l-1}$}
\nomenclature[13]{\(\bb_i\in\mathbb{R}\)}{Value $b_i$ augmented with terms from nullspace projections}
\nomenclature[14]{\({\lambda}\in\mathbb{R}^m\)}{Lagrange multipliers}
\nomenclature[15]{\(w\in\mathbb{R}^m\)}{Slack variables for inequality constraints}
\nomenclature[16]{\({\gamma}_{i}^+/\nu_{i}^+,{\gamma}_{i}^-/\nu_{i}^-\in\mathbb{R}\)}{Lagrange multipliers / slack variables for upper and lower auxiliary $\ell_0$ constraints; summarized by ${\gamma}_{i}^{\pm},\nu_{i}^{\pm}$}
\nomenclature[17]{\(\Psi_i,\Gamma_i\in\mathbb{R}\)}{Auxiliary variables $\Psi_i\coloneqq(w_i/\lambda_i)$ and $\Gamma_i\coloneqq(h_i/{{\gamma_i}})$}

\renewcommand{\nomname}{}
\vspace{-15pt}
\printnomenclature[1.8cm]
\normalsize

\section{Related Work}
\label{sec:relwork}

The concept of sparse programming originates from statistical analysis and decision-making problems such as compressed sensing~\citep{Candes2007} and portfolio optimization~\citep{wang2023}. In these domains, unconstrained linear regression problems are solved to fit model parameters to observed data. By introducing sparsity-promoting regularization, the optimization is biased toward selecting a minimal subset of parameters sufficient to explain the data. Sparsity is typically defined via the $\ell_0$-norm, which counts the number of non-zero entries in a vector $\chi$. Since the $\ell_0$-norm is a pseudo-norm (violating properties such as homogeneity, i.e., $\Vert \alpha \chi \Vert_{\ell_0} \neq \vert\alpha\vert\Vert \chi\Vert_{\ell_0}$ for $\alpha\neq 1$) its direct minimization requires a combinatorial search that becomes intractable for large-scale problems. To overcome this, the $\ell_0$-norm is often approximated by (weighted) $\ell_1$-norm minimization~\citep{Candes2007}, which can be efficiently solved using interior-point methods~\citep{Koh2007}.

In robotics, sparse control has been explored primarily through sequential optimization schemes. At each iteration, a sparse convex subproblem is solved by linearizing constraints related to robot dynamics and kinematics around the current operating point~\citep{pfeiffer2023}. In~\citep{Goncalves2015}, the sparsity-inducing properties of the simplex method are exploited to achieve sparse robot control with minimal joint activation via $\ell_1$-norm regularization. The methods proposed in~\citep{Polverini2019} and~\citep{Mingo2019} formulate sparse control and contact force selection as NP-hard mixed-integer linear programs, while in~\citep{Sathya2021} sparse $\ell_1$-norm terms are incorporated at multiple levels of a hierarchical controller. However, these approaches rely on $\ell_1$-norm relaxations and do not exploit true $\ell_0$-norm formulations, as considered here, which are required for versatile decision-making. %maximal sparsity in SHIK-C. {KP: not only for SHIK-C but also P}

Although the aforementioned methods enable sparse control, they lack nonlinear solver components such as Fletcher \textit{et al.}'s filter-based globalization technique~\citep{fletcher2002b}. This limits their applicability to nonlinear SHIK-P or sparse optimal control problems. Differential dynamic programming with linear $\ell_1$-norm regularization has been used to achieve sparse robot joint engagement~\citep{Dinev2021}; however, it cannot enforce sparsity in nonlinear cost functions. In~\citep{Hayashi2024}, $\ell_0$-norm regularization is used for the same purpose, but relies on a continuous surrogate function.
% but nonlinear cost functions remain dense and cannot enforce sparsity in the constraints. {KP: some of the original meaning and intention got lost here}
The optimization-based formulation in~\citep{Zhao2022} addresses nonlinear constrained sparse programming; however, nonlinear constraints are restricted to equalities, and sparsity is limited to constraints that are linear in the decision variables.

In summary, while sparse methods have been widely explored in control and estimation, their application to hierarchical, nonlinear robotic problems, particularly those requiring simultaneous autonomous location selection and IK reasoning, remains largely unexplored. This motivates the development of a dedicated nonlinear solver capable of handling sparse hierarchical programming with $\ell_0$-norm formulations.

\section{Sparse programming based decision-Making}
\label{sec:sparseProgDecMak}

Sparse programming can be applied to select a subset of feasible constraints from a group of similar-type ones for efficient \textit{decision-making}. In robotics, this concept naturally fits autonomous location selection, where a robot must identify a single feasible end-effector location from multiple potential candidates, e.g., from sensor detection or candidate grasp poses. We denote such a group of selection constraints as $\mathbb{S}\subseteq\mathbb{C}$. Each constraint in $\mathbb{S}$ is expressed as the following $\ell_2$-norm entry:
\begin{align}
	f_{s}(\chi_{\mathbb{S}}) = 
	\Vert g_{\mathbb{S}}(\chi_{\mathbb{S}}) - g_{d,s} \Vert_2^2,\qquad s\in\mathbb{S} 
	\label{eq:groupj}
\end{align}
The function $g_{\mathbb{S}}(\chi_{\mathbb{S}})\in\mathbb{R}^{m_g}$ is representative for all entries in $\mathbb{S}$ and depends on a subset of variables $\chi_{\mathbb{S}}\subseteq \chi$, e.g., the joint variables of a specific kinematic chain (e.g., right arm of a humanoid). The desired value $g_{d,s}\in\mathbb{R}^{m_g}$ is constant and can represent a Cartesian candidate location for the end-effector. 
Minimizing $\Vert (f_{ s})_{s\in\mathbb{S}}\Vert_{\ell_0}$ (with $(f_{ s})_{s\in\mathbb{S}}\coloneqq[ f_{s_1}\hspace{2pt}\dots]^T\in\mathbb{R}^{\vert{\mathbb{S}}\vert}$) with respect to the $\ell_0$-norm enables a unique selection --or \textit{decision-making}-- within the group $\mathbb{S}$, as formalized below.
\begin{prop}
	If all desired values $g_{d,s}$ are distinct and at least one constraint in $\mathbb{S}$ is feasible, then 
	$
	\min_{\chi_{\mathbb{S}}}\Vert (f_{s})_{s\in\mathbb{S}}\Vert_{\ell_0}= \vert\mathbb{S}\vert-1
	$
	is a global solution.
	\label{eq:uniquSel} 
\end{prop}
\begin{proof}
	From~\eqref{eq:groupj} and the given assumptions, only one entry $s$ can satisfy $f_{s}=0$ with $g_{\mathbb{S}} = g_{d,s}$.
\end{proof}

The identity $
\Vert 	(f_s)_{s\in\mathbb{S}} \Vert_{\ell_1} = 	\textstyle\sum_{s\in\mathbb{S}} \vert f_s\vert = \textstyle\sum_{s\in\mathbb{S}} f_s $
 shows that the $\ell_1$-norm does not affect the decision-functions of the form~\eqref{eq:groupj}, which emphasizes the versatility of the $\ell_0$-norm, as it is agnostic to their specific formulation.

\section{Sparse Hierarchical nonlinear Programming}
\label{sec:shnlp}

We introduce a novel optimization formalism for hierarchical decision-making coined SH-NLP. The formulation is rooted in optimization theory but directly corresponds to hierarchical robotic planning and control problems and forms the basis of the proposed sparse hierarchical inverse kinematics with autonomous location selection (Sec.~\ref{sec:hdm}):
\begin{align}
	\min_{\chi,(\nu_i)_{i\in\bC_l}}\qquad & \Vert (\nu_{i})_{i\in\bC_l}\Vert_{\ell_0}, \qquad l=1,\dots,p
	\label{eq:shnlp}\tag{SH-NLP}\\
	\text{s.t.}\qquad&
	f_{i}(\chi) \geqq \nu_{i}, \quad i\in\mathbb{C}_{l}\nonumber\\ 
	&f_{j}(\chi) \geq 0,\quad j\in\mathcal{I}_{\cup l-1}\nonumber\\
	 &f_{k}(\chi) = \nu_{k}^*,\quad k \in \mathcal{A}_{\cup l-1} \nonumber
\end{align}
The function $f_i(\chi)\in\mathbb{R}$ represents a nonlinear task constraint from the set $i\in\bC_l$, e.g., an end-effector position error. The function is parameterized by the variable vector $\chi\in\mathbb{R}^n$ (e.g., the robot’s joint state). The constraint set $\mathbb{C}_{l}= \{\bE_l,\bI_l\}$ includes equalities and inequalities, denoted by the symbol $\geqq$. The slack vector
$(\nu_{i})_{i\in\bC_l}$ allows controlled relaxation of these constraints, e.g., to handle unreachable position targets.
At each priority level $l$, the goal is to find the optimal feasible point $(\nu_i^*)_{i\in\bC_l}=0$ (all constraints satisfied) or the optimal infeasible point $(\nu_i^*)_{i\in\bC_l}\neq 0$ (with the fewest possible violations). Constraints that were active or inactive at previous levels ($1$ to $l-1$) must remain consistent, represented by the active and inactive sets $\mathcal{A}_{\cup l-1}$ and $\mathcal{I}_{\cup l-1}$, respectively. The symbol $\cup$ denotes the union over previous levels: $\mathcal{I}_{\cup l-1}\coloneqq \mathcal{I}_{1} \cup \cdots \cup \mathcal{I}_{l-1}$. Once $(\nu_i^*)_{i\in\bC_l}$ is identified, all equality and violated inequality constraints are added to the new active set $\mathcal{A}_{l}$ with corresponding optimal slack $(\nu_{i}^*)_{i\in\bC_{\cup l-1}}$, while satisfied inequalities form the inactive set $\mathcal{I}_l$. The process continues recursively for the next level $l \leftarrow l+1$. 

\textit{In the following, we omit the specification of set memberships $i\in\mathbb{C}_l$,	$s\in\mathbb{S}$, $ j\in\mathcal{I}_{\cup l-1}$, and $k \in \mathcal{A}_{\cup l-1}$ for brevity.}

Directly solving~\ref{eq:shnlp} is combinatorial and thus intractable for large-scale problems. Following the approach in~\citep{Candes2007} (originally designed for unconstrained linear programs), we adopt a continuous approximation of the $\ell_0$-norm minimization through a logarithmic surrogate, leading to:
\begin{align}
	\label{eq:l0approx}
	\min_{\chi,(\nu_i),(\tau_i)}\qquad& 
	\sum_{i\in\bC_l}\nolimits \log(\tau_{i} + \xi),
	\qquad l=1,\dots,p\\
	\text{s.t.}\qquad&
	-\tau_{i}\leq \nu_{i}\leq \tau_{i}\nonumber\\
	&
	f_{i}(\chi) \geqq \nu_{i}\nonumber\\ 
	&f_{j}(\chi) \geq 0\nonumber\\
	&  f_{k}(\chi) = \nu^*_{k} \nonumber
%	&\qquad i\in\mathbb{C}_{l},\quad j\in\mathcal{I}_{\cup l-1},\quad k \in \mathcal{A}_{\cup l-1} \nonumber
\end{align}
The summation of entry-wise logarithms, $\textstyle\sum\log(\cdot)$, rewards entries in $(\nu_i)$ with increasingly negative cost as they approach zero, thus promoting sparsity. The small constant $\xi>0$ ensures numerical stability. This emphasizes the benefit of prioritization, as weighting strategies are difficult to tune in this regime. The auxiliary variable $\tau_i$ and the corresponding bound constraints $\mathcal{T}_l$ on $\nu_i$ provide a smooth and continuous approximation of the discontinuous absolute value function $\vert \nu_i\vert$. This formulation maintains the hierarchical structure of the original sparse problem while enabling efficient numerical solutions through nonlinear programming.

%%%%
\section{Sequential Sparse Hierarchical Quadratic Programming}
\label{sec:shsqp}
\begin{figure*}[htp!]
	\centering
	\resizebox{2\columnwidth}{!}{%
		\begin{tikzpicture}[line cap=rect]
			\node[draw,line width=1pt,text width=5cm,fill=gray!20,rounded corners=3pt] (hlspsolve) at (-0.,-8.5) {
				{Solve \ref{eq:shqp} for ${x}_{\iota}$ with current trust-region radius $\rho_l$ by:
					\begin{itemize}
						\itemsep0pt
				\item {\textbf{\mbox{$\bm{\mathcal{N}}$\hspace{-2pt}QP} (Sec.~\ref{sec:ipm})}}
				\item standard QP solvers \citep{piqp,mosek}
					\end{itemize}
				}
			};
			
			\coordinate (b1)   at ($(hlspsolve.east) + (0.5,0)$);
			\coordinate (b2)   at ($(b1) + (0,1)$);
			\coordinate (b3)   at ($(b1) + (0,-1)$);
			
			\draw[line width=1pt] ($(hlspsolve.east)$) -- (b1);
			\draw[line width=1pt] (b1) -- (b2);
			\draw[line width=1pt] (b1) -- (b3);
			
			\node[] (inc) at ($(b1) + (1.35,1.55)$) {\color{Green}Planning (SHIK-P)};
			\node[] (inc) at ($(b1) + (1.3,-1.35)$) {\color{BrickRed}Control (SHIK-C)};
			
			\node[draw=Green,line width=1pt,text width=2.1cm,fill=gray!20,rounded corners=3pt] (convtest) at ($(b2)+(1.8,0.)$) {
				{Is $\Vert x_{\iota}\Vert_2 < \beta$?}
			};		
			
			\node[draw=Green,line width=1pt,text width=6.35cm,fill=gray!20,rounded corners=3pt] (filter) at ($(convtest)+(5.65,0.7)$) {
				{$\chi_{\iota+1},\rho_l\leftarrow$HSF$_l$($\chi_{\iota}+{x}_{\iota},\rho_l$); 
					\textbf{\citep{pfeiffer2024} adapted to \ref{eq:shnlp} as in Sec.~\ref{sec:shsqp}}}
			};
			
			\node[draw=Green,line width=1pt,text width=6.35cm,fill=gray!20,rounded corners=3pt] (conv) at ($(convtest)+(5.65,-0.7)$) {
	{\textbf{remove non-zero group constraints from $\mathbb{C}_{\bm{l}}$ per Sec.~\ref{sec:newtonm}}, 		 $(\nu_i^*)_{i\in\bC_l}=(\nu_i)_{i\in\bC_l}$		 , $(\omega_{i}^*)_{i\in\bC_l}=(\omega_{i})_{i\in\bC_l}$
}
	};
			
	\node[draw=Green,line width=1pt,text width=1.8cm,fill=gray!20,rounded corners=3pt] (incl) at ($(conv)+(5.,0)$) {
				{$l$++\\if $l = p+1$:\\\textit{exit}}
			};

			\coordinate (b15)   at ($(convtest.east) + (0.4,0)$);
			\coordinate (b16)   at ($(b15) + (0,0.7)$);
			\coordinate (b17)   at ($(b15) + (0,-0.7)$);
			
			\coordinate (b18)   at ($(filter.east) + (3.225,-0)$);
			\coordinate (b19)   at ($(incl.east) + (0.5,-0)$);	
			\coordinate (b20)   at ($(incl.east) + (0.5,-2)$);	
			\coordinate (b21)   at ($(b20) + (-17.096,-0.)$);			
			
			\draw[Green,line width=1pt,decoration={markings,mark=at position 1 with
				{\arrow[scale=2,>=latex]{>}}},postaction={decorate}] ($(b2)$) -- (convtest.west);
			
			\draw[Green,line width=1pt] ($(convtest.east)$) -- (b15);
			\draw[Green,line width=1pt] (b15) -- (b16);
			\draw[Green,line width=1pt] (b15) -- (b17);
			
			\draw[Green,line width=1pt,decoration={markings,mark=at position 1 with
				{\arrow[scale=2,>=latex]{>}}},postaction={decorate}] (b16) -- (filter.west);
			\draw[Green,line width=1pt,decoration={markings,mark=at position 1 with
				{\arrow[scale=2,>=latex]{>}}},postaction={decorate}] (b17) -- (conv.west);
			\draw[Green,line width=1pt,decoration={markings,mark=at position 1 with
				{\arrow[scale=2,>=latex]{>}}},postaction={decorate}] (conv.east) -- (incl.west);
			
			\draw[Green,line width=1pt] ($(filter.east)$) -- (b18);
			\draw[Green,line width=1pt] ($(incl.east)$) -- (b19);
			\draw[line width=1pt] (b18) -- (b20);
			\draw[line width=1pt] (b20) -- (b21);
			
			\draw[line width=1pt,decoration={markings,mark=at position 1 with
				{\arrow[scale=2,>=latex]{>}}},postaction={decorate}] (b21) -- ($(hlspsolve.south)$);
			
			\draw[line width=1pt,decoration={markings,mark=at position 1 with
				{\arrow[scale=2,>=latex]{>}}},postaction={decorate}] ($(hlspsolve.north) + (0.0,0.7)$) -- ($(hlspsolve.north)$);
			
			\node[] (inc) at ($(hlspsolve.north) + (0.,0.95)$) { 
				$\iota=0$, $l=1$, $\rho_l=\rho_0$, $\chi = \chi_0$
			};
			\node[] (inc) at ($(b21) + (1.2,0.25)$) { 
				$\iota$++, $\rho_l=\rho_0$
			};
			\node[] (inc) at ($(convtest) + (1.75,0.95)$) { No };
			\node[] (inc) at ($(convtest) + (1.75,-0.95)$) { Yes };
			
			\node[draw=BrickRed,line width=1pt,text width=5cm,fill=gray!20,rounded corners=3pt] (realtime) at ($(b3)+(7,0.)$) {
				{$\chi_{\iota+1}\leftarrow \chi_{\iota}+{x}_{\iota}$, $\rho_l$ constant~\citep{pfeiffer2023}}};
			
			\draw[BrickRed,line width=1pt,decoration={markings,mark=at position 1 with
				{\arrow[scale=2,>=latex]{>}}},postaction={decorate}] (b3) -- ($(realtime.west)$);
			
			\draw[BrickRed,line width=1pt] ($(realtime.east)$) -- ($(realtime.east) + (4.35,0)$);
			
	\end{tikzpicture}}
	\caption{Symbolic overview of the sequential sparse hierarchical quadratic programming (S-SHQP) algorithm with trust region and hierarchical step-filter (HSF)~\citep{pfeiffer2024}, adapted from the SQP step-filter of Fletcher \textit{et al.}~\citep{fletcher2002b}. The framework efficiently solves sparse hierarchical nonlinear programs~\eqref{eq:shnlp} across $p$ priority levels. New contributions are printed in bold.}
	\label{fig:scheme}
\end{figure*}

We solve~\eqref{eq:l0approx} by our nonlinear sparse solver S-SHQP, which is based on the principles of sequential quadratic programming (see~\citep{nocedal2006}, Ch.~18). As depicted in Fig.~\ref{fig:scheme}, at each iteration $\iota$, a SHQP subproblem approximates~\eqref{eq:l0approx} around the current state $\BIN \chi_{\iota}^T & \nu_{\iota}^T & \tau_{\iota}^T\BOUT^T$, which is solved for the local step $\BIN {x}_\iota^T & v_\iota^T & t_{\iota}^T\BOUT^T$~\citep{nocedal2006,Candes2007,pfeiffer2023}.  
SHQP results from the first-order conditions of~\eqref{eq:l0approx} when Newton’s method is applied to them:
\begin{align}
	\min_{{z_l},(v_i),({t_i})} \quad& \textstyle\sum_{i\in\bC_l}\omega_{i}{t}_{i}, \quad l=1,\dots,p\label{eq:shqp}\tag{SHQP}\\[-7pt]
	&\colorbox{gray!20}{$\displaystyle{+ 0.5}(x_{l-1}^{*} + N_{l-1}z_l)^T H_l (x_{l-1}^{*} + N_{l-1}{z}_l)$}\nonumber\\
	\text{s.t.}\qquad& -{t}_{i} \leq {v}_{i} \leq {t}_{i}\nonumber\\
	 &\ta_{i}^T{{z}_l} - \bb_{i} \geqq {v}_{i}\nonumber\\
	&\ta_{j}^T{{z}_l} - \bb_{j}\geq 0\nonumber\\
		&	\Vert x_l\Vert_{\infty} < \rho\nonumber
\end{align}
The weights $\omega_i>0$ are defined as:
\begin{equation}
	\label{eq:omega}
	\omega_{i} =  1/{{t}_{i} + \xi}
\end{equation}
Setting $(\omega_{i})={1}$ recovers the $\ell_1$-norm case of~\ref{eq:shnlp}.

The change of variables $x_l = x_{l-1}^* + N_{l-1}z_l$ (with $x\coloneqq x_p$) represents a projection into the nullspace of the active constraints $\mathcal{A}_{\cup l-1}$. Here, $x_{l-1}^*$ is the optimal primal of levels $1$ to $l-1$, and $N_{l-1}\in\mathbb{R}^{n\times n_r}$ is a nullspace basis of the vectors $a_k\in\mathbb{R}^n$ ($a_{k}^TN_{l-1}=0$; $N_0=I$). The vector ${z}_l\in\mathbb{R}^{n_r}$ is the projected primal of level $l$. This projection eliminates both active constraints and already occupied variables such that the number of remaining variables is $n_r<n$.

The hierarchical Lagrangian Hessian ${H}_l$~\citep{pfeiffer2023} captures second-order terms of the constraints, and is crucial for ensuring linear constraint qualifications (LCQ) of the vectors $a_i=\mathrm{grad} f_i(\chi_{\iota})$~\citep{fletcher2002b} (with projected counterparts  $\tilde{H}_l = N_{l-1}^TH_lN_{l-1}$ and $\ta_i^T = a_i^TN_{\l-1}$).
The term is highlighted in gray to emphasize the difference between Gauss-Newton (no $H_l$) and full Newton steps (with $H_l$). The choice between the two depends on linearized constraint feasibility $\Vert({v}_{i}^*)\Vert_2 \leq \hspace{-4pt}/ > \epsilon$~\citep{pfeiffer2018}, where $\epsilon>0$ is a numerical threshold. In practice, Newton's method is needed in singular robot tasks in the presence of unreachable targets. 

The value $b_i$ is defined as $b_i \coloneqq f_i(\chi_{\iota})$, and $\breve{b}_i$ denotes the nullspace projection offset from previous priority levels: $\breve{b}_{i} \coloneqq b_{i} - a_{i}^Tx_{l-1}^*$.
The primal $x_l$ is restricted within a trust region of radius $\rho$, which enforces local validity of the SHQP model at $\chi_{\iota}$ (omitted hereon for brevity). 
S-SHQP differs from NL-HLSP, which solves a hierarchical least-squares problem similar  to~\ref{eq:shqp}, but with quadratic cost $\Vert (v_i)\Vert_2^2$  and no auxiliary constraints $\mathcal{T}_l$.

Depending on the operational mode, the algorithm proceeds as follows:

\textit{1) Planning:}
The proposed step is accepted or rejected by the HSF~\citep{fletcher2002b,pfeiffer2024}, which evaluates both feasibility and optimality. The filter is updated with the values $\textstyle\sum_{i\in\mathbb{C}_l}\log(|f_{i}^{\geq0}|+\epsilon)$  for each level $l$, requiring new points to strictly improve over prior ones. The expression $f_{i}^{\geq0}\coloneqq  \max(0,f_{i}^T)$ maintains positive values of the inequality constraints $i\in\bI_l$. Model validity is checked by comparing the nonlinear values $f_{i}^{\geq0}$ to the linearized approximations $\omega_{i}(a_i^Tx-b_{i})^{\geq0}$. The trust-region radius is then adapted (expanded for accepted steps, reduced otherwise). Once the S-SHQP has converged with $\Vert{x}_{\iota}\Vert_2 < \beta$ ($\beta > 0$), the optimal slacks are stored. Sparse constraints $s$ with $\nu_{s}^*=0$ in group $s\in\mathbb{S}$ (Sec.~\ref{sec:hdm}) trigger removal of non-sparse ones $\nu_{s}\neq 0$ from $\mathbb{C}_l$. Corresponding weights are fixed as $\omega^*_{s} = \omega_{s,\iota}$, while $\omega_{s}$ is continuously updated according to~\eqref{eq:omega} for unsolved levels ($l+1$ to $p$). Feasibility of~\eqref{eq:l0approx} is guaranteed only at full HSF convergence.

\textit{2) Control:}
Each step is directly accepted. With a properly tuned constant trust-region radius, the resulting state remains approximately feasible for~\eqref{eq:l0approx}, which makes the update suitable for closed-loop robotic control~\citep{pfeiffer2023}.

\section{Parallel Hierarchical Decision-Making}
\label{sec:hdm}

\subsection{Avoiding Double Allocation in parallel decision-making}

In robotics, it may be necessary for several selection groups $\mathbb{S}_{c=1,2,...}\subset\mathbb{S}$ to choose from the same set of potential candidates. To avoid \emph{double allocation}, e.g., when both robot feet or arms attempt to select the same candidate, the following modified version of the weights $({\omega}_{s_c})$
with $s_c\in\mathbb{S}_c$ is used:
\begin{align}
	(\tilde{\omega}_{s_c}) = 
	(\phi_{s_c})\odot({\omega}_{s_c}), 
	\hspace{5pt}
	(\phi_{s_c}) = 
	\min((f_{s_c}))\,\vert (f_{s_c})\vert^{-1}
	\label{eq:l0mask}
\end{align}
The symbol $\odot$ symbolizes the entry-wise multiplication between two vectors. The operation $\vert(f_{s_c})\vert^{-1}$ returns the reciprocal of each individual entry of the vector.
The masks $\phi_{s_c}$ are computed in sequence and represent the minimum value of the constraints in each set $\mathbb{S}_c$. In order to prevent multiple groups from converging to the same candidate, each minimum index of each set $\mathbb{S}_c$ is chosen uniquely over all sets, while already identified indices from other sets are set to zero in $(\phi_{s_c})$.

Since in each S-SHQP iteration the mask $(\phi_{\mathbb{S}_{\mathbb{C}}})$ is treated as constant, the resulting \ref{eq:shqp} subproblem becomes a less accurate local approximation of the original problem~\eqref{eq:l0approx}. However, the HSF robustly compensates for this, and no adverse effects on convergence were observed in practice.

\subsection{Implications for Newton’s Method and Hierarchical Decision-making}
\label{sec:newtonm}

In the case of Newton’s method (see Sec.~\ref{sec:shsqp}), the Lagrangian Hessian is non-zero and full-rank over the variables $\chi_{\mathbb{S}}$ of a group $\mathbb{S}$. This guarantees solver convergence via LCQ, but also prevents reuse of these variables in lower-priority levels. As a result, an unnecessary switch of Newton’s method can degrade accuracy on lower priority levels in \textit{hierarchical decision-making}. To mitigate this, we leverage feasible entries within $\mathbb{S}$ whenever available.
\begin{itemize}
	\item Run the HSF for level $l$ and identify the optimally infeasible point $(\nu_{s}^*)$ of group $\mathbb{S}$.
	\item If a feasible constraint $s$ exists (i.e., $\nu_{s} \leq \zeta$, $\zeta > 0$), add it to the active set $\mathcal{A}_l$ and discard all infeasible ones.
	\item If no feasible constraint exists, add one representative constraint from $\mathbb{S}$ to $\mathcal{A}_l$ and discard the others.
\end{itemize}
Discarding constraints is justified by Theorem~\ref{eq:uniquSel}: once a single constraint $s$ of $\mathbb{S}$ is included in the active set, all other slacks in $\mathbb{S}$ are uniquely determined. Specifically, if $\Vert g_{\mathbb{S}}(\chi_{\mathbb{S}}) - g_{d,s_1}\Vert_2^2 = \nu_{s_1}$ holds for some $\chi_{\mathbb{S}}$ and $s_1\in\mathbb{S}$, then $\Vert g_{\mathbb{S}}(\chi_{\mathbb{S}}) - g_{d,s_2}\Vert_2^2 = \nu_{s_2}$ is implicitly defined for any other $s,s_2\in\mathbb{S}$.  
Selecting the zero constraint $s_1$ (the feasible one) avoids activating Newton’s method since the switching condition $\Vert{\nu}_{s_1}^*\Vert \not> \epsilon$ remains unsatisfied.

%%%
\section{An Interior-Point Method for SHQP}
\label{sec:ipm}

Efficient solvers have been proposed to solve HLSP as sub-problems of NL-HLSP, using either active-set~\citep{escande2014} or interior-point methods~\citep{pfeiffer2021}. In these hierarchical formulations, each level $l$ is solved sequentially, and its least-squares objective becomes a linear constraint for the lower levels $l+1$ to $p$. However, this cascaded scheme cannot be directly applied to~\ref{eq:shqp}, since its objective includes the linear term $\sum_{i\in\mathbb{C}_l}\omega_{i}{t}_{i}$, which prevents reformulation into a least-squares problem. This limitation can be resolved by the following observation:
\begin{theorem}
	\label{th:act}
	${v}_{i}$ is strictly upper or lower bounded at ${t}_{i}$ or $-{t}_{i}$ for $\omega_{i} > 0$ and the infeasible case $v_{i}\neq 0$. Otherwise ${t}_{i} = v_{i} = 0$.
\end{theorem}

As a result, the inactive opposing constraint will never be active and can be omitted in $\mI_{\cup l-1}$. For example, if ${t}_{i} - {v}_{i}=0$ holds for constraint $i$, then ${t}_{i} + {v}_{i} \geq 0$ always holds. Furthermore, this allows one not only to store the optimal slack ${v}_i^*$, but also to set the optimal auxiliary variable ${t}_i^* = \vert{v}_i^*\vert$. Since the term $\omega_{i}{t}_{i}^*$ in the cost function is now constant, it can be omitted.
The cost function of each level is then a least-squares problem ($\Vert R_l( x_{l-1}^* + N_{l-1}{{z}_l}) \Vert_2^2$) and becomes the linear constraint $R_lN_{l-1}{{z}_l} = R_lx_{l-1}^*$ for all subsequent levels $l+1$ to $p$. Here, ${R}_l$ is a factor of the semi-positive definite Hessian such that ${H}_l = {R}_l^T{R}_l$. This reformulation enables the design of an efficient interior-point solver, denoted as \mbox{$\mathcal{N}$\hspace{-2pt}QP}. The proof of Theorem~\ref{th:act} follows after some preliminaries of the solver have been outlined.

In the following, the slack variables $h_{i}^{\pm}\coloneqq[ h_{i}^{+} \hspace{3pt} h_{i}^{-}]^T$, $w_{i}$ (for $i\in\mathbb{I}_l$ only, highlighted with gray crosshatch pattern), and $w_{j}$ are introduced for the different inequality constraints in \ref{eq:shqp}. These are penalized using a log-barrier function with centering parameter $\sigma$ and duality measure $\mu$~\citep{pfeiffer2021}. This leads to the following problem formulation:
\begin{align}
	\label{eq:ipmqp}
	&\min_{{z},({v}_i),({t}_{i}),(h_i^{\pm}),(w_{i/j})} \quad \textstyle \sum_{i\in\mathbb{C}_l}\omega_{i}{t}_{i} \qquad l=1,\dots,p\\[0pt]
	&\qquad\colorbox{gray!20}{$\textstyle{+ 0.5}(x_{l-1}^{*} + N_{l-1}z_l)^T H_l (x_{l-1}^{*} + N_{l-1}{z}_l)$}\nonumber\\
	&\hspace{-15pt}- \hspace{-1pt}\sigma\mu\hspace{-1pt}\left(\hspace{-0pt}
	\textstyle \sum_{i\in\mathbb{C}_l}\hspace{-1pt}\log(h_{i}^{\pm})  \hspace{-1pt}+\hspace{-1pt}\textstyle \sum_{i\in\mathbb{I}_l}\hspace{-1pt}\log(w_{i}) 
	\hspace{-1pt}+\hspace{-1pt}\textstyle \sum_{j\in\mI_{\cup l-1}}\hspace{-2pt}\log(w_{j})\hspace{-1pt}\right) \nonumber\\
	\text{s.t.}
	\quad&  {t}_{i} -  v_{i} =  h_{i}^+,\qquad {t}_{i} + v_{i} = h_{i}^- \nonumber\\
	\quad& \ta_{i}^T{z}_l - \bb_{i} = v_{i} \stripedbox{$\displaystyle{+ w_{i}}$} \nonumber\\
	& \ta_{j}^T{z}_l - \bb_{j} = w_{j}\nonumber\\
	& h_{i}^{\pm}\geq 0, \quad \stripedbox{$\displaystyle{w_{i}\geq 0}$}, \quad w_{j} \geq 0 \label{eq:compcond}
\end{align}
Variables with superscripts $\pm$ indicate upper and lower bounds in the constraint set $\mathcal{T}_{l}$.

The corresponding Lagrangian of~\eqref{eq:ipmqp} is given by:
\begin{align}
	&\mathcal{L}_l \coloneqq \textstyle \sum_{i\in\mathbb{C}_l}\omega_{i}{t}_{i} + 0.5(x_{l-1}^* + N_{l-1}{{z}_l})^T {H}_l (x_{l-1}^* + N_{l-1}{{z}_l})\nonumber\\
	&- \hspace{-1pt}\sigma\mu\hspace{-1pt}\left(\hspace{-0pt}
	\textstyle \sum_{i\in\mathbb{C}_l}\hspace{-1pt}\log(h_{i}^{\pm})  \hspace{-1pt}+\hspace{-1pt}\textstyle \sum_{i\in\mathbb{I}_l}\hspace{-1pt}\log(w_{i}) 
	\hspace{-1pt}+\hspace{-1pt}\textstyle \sum_{j\in\mI_{\cup l-1}}\hspace{-2pt}\log(w_{j})\hspace{-1pt}\right)\nonumber\\
	& - \textstyle \sum_{i\in\mathbb{C}_l}{\gamma}_{i}^+ ( {t}_{i}\hspace{-0pt}-\hspace{-0pt}v_{i} \hspace{-0pt}-\hspace{-0pt}h_{i}^+) + \textstyle \sum_{i\in\mathbb{C}_l}
	{\gamma}_{i}^-( {t}_{i} \hspace{-0pt}+\hspace{-0pt} v_{i} \hspace{-0pt}-\hspace{-0pt} h_{i}^-)\nonumber\\
	& -\textstyle \sum_{i\in\mathbb{C}_l} \lambda_{i}\left(\ta_{i}^T{z}_l - \bb_{i} - v_{i} \stripedbox{$\displaystyle{- w_{i}}$}\right)\nonumber\\
	& - \textstyle \sum_{j\in\mI_{\cup l-1}}\lambda_{j}(\ta_{j}^T{z}_l - \bb_{j} - w_{j})
\end{align}
$\gamma$ and ${\lambda}$ are the Lagrange multipliers associated with the constraints $\mathcal{T}_l$, $\mathbb{C}_l$, and  $\mI_{\cup l-1}$.
We summarize the problem variables as:
\begin{align}
	q_l = 
	[&	{z}_l^T \hspace{1pt} 
	(v_{i})^T \hspace{2pt} 
	({t}_{i})^T \hspace{2pt}
	(\gamma_{i}^{\pm})^{T} \hspace{1pt} 
	({\lambda}_{i})^T \hspace{2pt} 
	({\lambda}_{j})^T \hspace{2pt} 
	 (h_{i}^{\pm })^{T} \hspace{1pt} 
	\stripedbox{$\displaystyle{(w_{i})}$}^T \hspace{0pt}
 (w_{j})^{T} ]^T
\end{align}

The first-order optimality, or Karush-Kuhn-Tucker (KKT), conditions are expressed as:
\begin{equation}
	K_{q_l}\coloneqq\nabla_{q_l} \mathcal{L}_l(q_l) = 0 \label{eq:kkt}\tag{KKT}
\end{equation}

The components of the Lagrangian gradient  $K_l(q_l)$ are
\begin{align}
	&K_{{z_l}} 	\hspace{-0pt}=	\hspace{-0pt} \tilde{{H}}_l{z}_l	\hspace{-0pt} +	\hspace{-0pt} N_{l-1}^T {H}_lx_l    
	\hspace{-0pt}-	\hspace{-3pt}\textstyle \sum_{i\in\mathbb{C}_l} \ta_{i}{\lambda}_{i} 		\hspace{-0pt} - 	\hspace{-3pt}\textstyle \sum_{j\in\mI_{\cup l-1}} \ta_{j}{\lambda}_{j} \\
	&K_{{\lambda}_{i}} \hspace{-0pt}=  	-\ta_{i}^T{z}_l + b_{i} +  {v}_{i} \stripedbox{$\displaystyle{+  w_{i}}$}\\
	&K_{{\lambda}_{j}} \hspace{-0pt}= -\ta_{j}^Tz_l + b_{j} + w_{j} \\
	&K_{{v}_{i}} = 	{\gamma}_{i}^+ \hspace{-0pt} - \hspace{-0pt} {\gamma}_{i}^- \hspace{-0pt}+\hspace{-0pt} {\lambda}_{i}\\
	&  K_{{t}_{i}} = 	\omega_{i} \hspace{-0pt}-\hspace{-0pt} {\gamma}_{i}^+ \hspace{-0pt}-\hspace{-0pt} {\gamma}_{i}^- \\
	&K_{{\gamma}_{i}^{+}} \hspace{-0pt}=\hspace{-0pt}  {t}_{i} \hspace{-0pt}-\hspace{-0pt} {v}_{i}  \hspace{-0pt}-\hspace{-0pt} h_{i}^+,\qquad K_{{\gamma}_{i}^{-}} \hspace{-0pt}= {t}_{i} \hspace{-0pt}+\hspace{-0pt} {v}_{i}  \hspace{-0pt}-\hspace{-0pt} h_{i}^- \\
	& K_{w_{i/j}} \hspace{-0pt}=\hspace{-0pt} {\lambda}_{i/j} w_{i/j} \hspace{-0pt}-\hspace{-0pt} \sigma\mu ,\qquad
	K_{h_{i}^{\pm}} \hspace{-0pt}=\hspace{-0pt} {\gamma}_{i}^{\pm} h_{i}^{\pm} \hspace{-0pt}-\hspace{-0pt} \sigma\mu 
\end{align}

With this foundation, the proof of Theorem~\ref{th:act} is given:
\begin{proof}
	We prove Theorem~\ref{th:act} by contradiction, considering the~\ref{eq:kkt} conditions of the~\ref{eq:shqp}. 	
	If both the lower and upper bounds of the auxiliary constraint are inactive, we obtain
	${\gamma}_{i}^+ = {\gamma}_{i}^- = 0$.
	This leads to a conflict between the conditions 
	$K_{{t}_{i}} = \omega_{i} = 0$ 
	and $\omega_{i} > 0$.  
	Now, consider the case of double-sided activity, 
	\mbox{${\gamma}_{i}^+, {\gamma}_{i}^- > 0$}.
	Due to complementary slackness, we must have  
	$h_{i}^+ = h_{i}^- = 0$.
	This contradicts the condition 
	$K_{{\gamma}_{i}^{\pm}} = 0$, 
	which yields 
	${t}_{i} = v_{i}$ 
	and 
	${t}_{\mathbb{C}_l} = -v_{\mathbb{C}_l}$.
	This can only hold in the feasible case 
	$v_{i} = {t}_{i} = 0$.  
	Therefore, necessarily, only one bound of each constraint pair 
	$-{t}_{i} \leq {v}_{i} \leq {t}_{i}$ 
	is active whenever $v_{i} \neq 0$.
\end{proof}

Applying Newton’s method to the nonlinear KKT system yields the linearized update:
\begin{align}
	K_{q_l}(q_l+\Delta q_l) &\approx K_{q_l}(q_l) + \nabla_{q_l}K_{q_l}(q_l)\Delta q_l = 0
\end{align}
Following variable substitutions can be obtained. 
We define $\Psi_i\coloneqq w_i/{\lambda_i}$ and $\Gamma_{i}^{\pm}=h_{i}^{\pm}/{\gamma}_{i}^{\pm}$.
\begin{align}
	&\Delta\gamma_{{i}}^-=
	1/(\stripedbox{$\displaystyle{4\Psi_{i}}$} + \Gamma_{i}^{+} + \Gamma_{i}^{-})(-2a_{i}^T\Delta z \\	
	& -2(
	\stripedbox{$\displaystyle{- \Psi_{i} ( K_{t_{i}}  + K_{v_{i}}) - K_{w_{i}}/\lambda_{{i}}}$}
	 +K_{\lambda_{i}}) \nonumber\\
	&-(-\Psi_{t_{i}}^{+} K_{t_{i}} - K_{h_{{i}}^+}/\gamma_{{i}}^{+}	-K_{\gamma_{i}^+}) - K_{h_i^-}/\gamma_{{i}}^{-}	-K_{\gamma_{i}^-})\nonumber\\
	&\Delta \gamma_{i}^+ =  -\Delta\gamma_{i}^- + K_{t_{i}}\\
	&\Delta \lambda_{i} =  \Delta \gamma_{i}^+ - \Delta \gamma_{i}^- + K_{v_{i}}\\
	&\Delta \lambda_{j}   = 1/\Psi_{j}(a_{j}^T\Delta z - K_{w_{{j}}}/\lambda_{j} +K_{\lambda_{j}})\\
	&\Delta w_{i/j} = -(w_{i/j} \Delta \lambda_{i/j} + K_{w_{i/j}}) / \lambda_{i/j} \\
	&\Delta h_i^{\pm} = -(h_{i}^{\pm} \Delta \gamma_{i}^{\pm} + K_{h_{i}^{\pm}}) / \gamma_{i}^{\pm} \\
	&\Delta v_{i} = a_i^T\Delta z_l + \stripedbox{$\displaystyle{\Delta w_{i} }$} + K_{\lambda_{i}}\\
	& \Delta t_{i} =  \Delta v_{i}  +  \Delta h_{i}^+ - K_{\gamma_{i}^-}
\end{align}
This leads to the main computational step of the interior-point method:
\begin{align}
	&(\tilde{H}_l + \textstyle \sum_{i\in\mathbb{C}_l}C_{i} + \textstyle \sum_{j\in\mI_{\cup l-1}} C_j )\Delta{z}_l	\label{eq:newtonstep} \\
	=& - K_{{z}} + \textstyle \sum_{i\in\mathbb{C}_l}r_i + \textstyle\sum_{j\in\mI_{\cup l-1}}r_{j} \nonumber
\end{align}
The left-hand side matrices $C$ are defined below: 
\begin{align}
	C_{i} &\coloneqq  4/( \Gamma_{i}^{+} + \Gamma_{i}^{-} \stripedbox{$\displaystyle{+{4\Psi_{i}}}$})\ta_{i}\ta_{i}^T \label{eq:lhs1}\\
	C_{j} &\coloneqq 1/\Psi_{j}\ta_{j}\ta_{j}^T\label{eq:lhs2}
\end{align}
The corresponding right-hand side vectors $r$ are given by:
\begin{align}	
	&r_{i} \coloneqq
	\ta_{i}(2/ (\stripedbox{$\displaystyle{4\Psi_{i}}$} + \Gamma_{{i}}^{+} + \Gamma_{i}^{-})(\Gamma_{i}^{+} K_{{t}_{i}}\\
	&+ K_{h_{{i}}^+}/{\gamma}_{i}^{+}  
+2(K_{\lambda_i}
	 \stripedbox{$\displaystyle-{\Psi_i} 
		(-
		K_{t_i} -
		K_{v_i}		) -
	K_{w_{i}}/	\lambda_i $}) \nonumber\\
	&- K_{h_{i}^-}/{\gamma}_{i} 
	+K_{\gamma_{i}^+} 	-K_{\gamma_i^-})-  K_{t_i}\ - K_{v_{i}})\nonumber
	\\
	&r_{j} \coloneqq  \ta_{j} (- K_{w_{j}}/{\lambda}_{j} +K_{{\lambda}_{j}})/\Psi_{j}
\end{align}

Computing the Newton step $\Delta {z}_l$ in~\eqref{eq:newtonstep} requires $O(n_r^3)$ operations, for example when using a Cholesky decomposition. 
The number of sparse constraints $\vert{\bC_l}\vert$ affects the matrix-matrix multiplications in~\eqref{eq:newtonstep}  only linearly ($O(n_r^2\vert{\bC_l}\vert)$). 
This scaling behavior aligns with results from unconstrained $\ell_1$-regularization methods~\citep{Koh2007,Loris2009,Borsic2012}, but contrasts with common robotics formulations~\citep{Berret2008,Polverini2019}, which exhibit cubic scaling $O((n+\vert{\bC_l}\vert)^3)$ because auxiliary variables are not eliminated.

The Newton step in~\eqref{eq:newtonstep} is then iteratively computed and added to the current primal variable ${z}_l$, using a line-search factor to maintain dual feasibility~\eqref{eq:compcond} and the condition 	\mbox{${\gamma}^{\pm},{\lambda} \geq 0$}.
Once the nonlinear KKT conditions converge ($K_q(q_l)\approx0$), the new active and inactive constraint sets $\mathcal{A}_{l}$ and $\mathcal{I}_{l}$ are assembled. 
The constraint matrices of the remaining hierarchy levels are then projected into the nullspace of the new active set. 
This process is repeated across all $p$ levels. 
Further details on interior-point formulations for hierarchical optimization can be found in~\citep{pfeiffer2021}.

\section{Evaluation}
\label{sec:eval}

We evaluate the proposed S-SHQP framework on both hierarchies of numerical test functions and robotic applications, including \mbox{SHIK-P} and \mbox{SHIK-C}.  
In each~\ref{eq:shqp}, a trust-region constraint is defined at the zeroth level. Variable regularization or nonlinear robotics constraints like collision avoidance are defined in the $\ell_2$-norm and can be solved as described in~\citep{pfeiffer2024}. 
The nullspace basis of active constraints is computed following~\citep{pfeiffer2021}.

All algorithms are implemented in \texttt{C++} using the \texttt{Eigen} library~\citep{eigenweb}. 
In the robotic examples, the SH-NLP and~\ref{eq:shqp} sub-problems are computed using \texttt{Pinocchio}~\citep{carpentier2019pinocchio}. 
We compare our proposed \mbox{$\mathcal{N}$\hspace{-2pt}QP} solver to state-of-the-art QP solvers \mbox{(H-)MOSEK}~\citep{mosek} and \mbox{(H-)PIQP}~\citep{piqp} (H: hierarchical). 
Theorem~\ref{th:act} is applied to all solvers so that auxiliary variables and constraints are not propagated to lower-priority levels. 
Simulations are executed on an Intel Core Ultra 9 185H$\times$22 CPU with 64~GB RAM.

First, a pick-and-place scenario with a manipulator as a simplified version of SHIK-P is used for comparison with state-of-the-art nonlinear programming (NLP) and MINLP solvers (Sec.~\ref{sec:shikpxarm6}).
S-SHQP is then verified on a hierarchy of typical test functions to evaluate convergence in hierarchical decision-making (Sec.~\ref{sec:eval:testfunc}). 
Efficient \mbox{SHIK-P} with many candidate solutions is validated in humanoid planning for autonomous selection among many potential and prioritized end-effector locations (Sec.~\ref{sec:eval:plan}). 
Real-time \mbox{SHIK-C} for a manipulator tracking task with sparse joint activation is presented in Sec.~\ref{sec:eval:realtime}. 
Efficient \mbox{SHIK-C} with a large-scale candidate sets is evaluated with a humanoid (Sec.~\ref{sec:eval:continSelect}). Potential robotic applications are described in Sec.~\ref{sec:robotics}. Real-world robot experiments on UFactory's \texttt{xarm6} are given for Sec.~\ref{sec:shikpxarm6} and Sec.~\ref{sec:eval:realtime} (see video).

\subsection{SHIK-P of manipulator for pick-and-place}
\label{sec:shikpxarm6}
\begin{table}[htp!]
	\centering
	\resizebox{0.8\columnwidth}{!}	{%
		\begin{tabular}{@{} cccccc @{}}  
			\toprule
		$\underline{l}$ & & $\ell$ & $f_i(\chi)\leqq \nu_i,\quad i\in\bC_l$ \\
		\midrule
		$\underline{1}$ & Joint ang. lim.& 2 & {$\underline{\chi} \leq \chi \leq \overline{\chi}$}   \\
		& Coll. av. & 2 & $f_{\text{coll}}(\chi) \geq 0$   \\
		\rowcolor{gray!20}
		\underline{2} & {EF} & 0 / 1 / 2  & $\Vert f_{\text{EF}}(\chi) - f_{\text{EF},d,i}\Vert_2^2 = \nu_{i}$\\
			\bottomrule
		\end{tabular}
	}
	\caption{\mbox{SHIK-P} for pick-and-place of $\vert\mathbb{S}\vert=10,...,100$ objects.}
	\label{tab:xarm6shikp}
\end{table}

\begin{figure}[htp!]
	\includegraphics[width=1 \columnwidth]{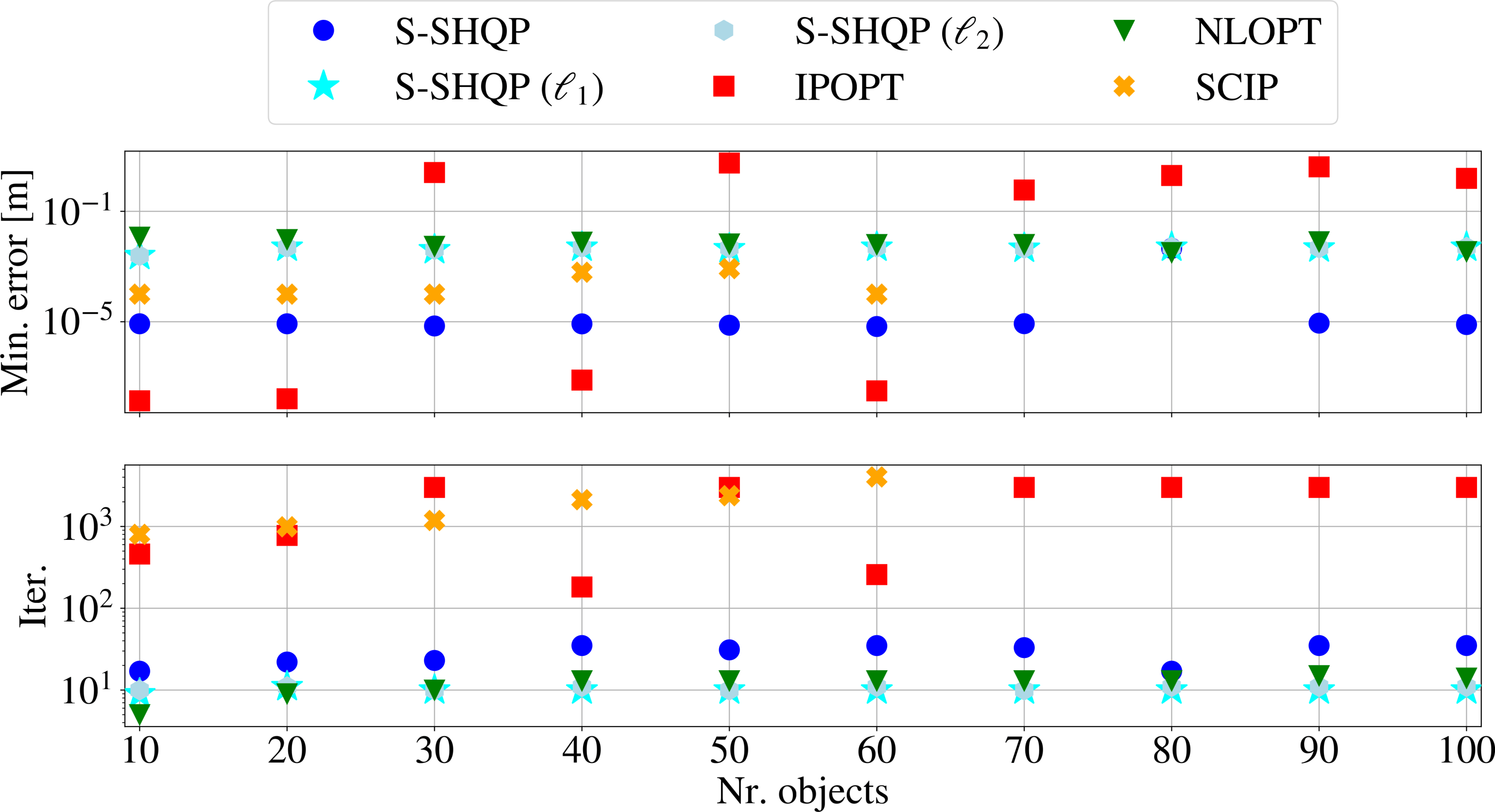}
	\centering
	\caption{SHIK-P for pick-and-place: number of nonlinear solver iterations (Iter.) and minimum decision-making error when picking one out of 10 to 100 objects.}
	\label{fig:xarm6shikp}
\end{figure}
To the best of our knowledge, there is currently no other solver except S-SHQP which can solve the approximation~\eqref{eq:l0approx} of~\ref{eq:shnlp}. In order to benchmark our method nonetheless, we compare it with the nonlinear solvers IPOPT~\citep{ipopt} and NLOPT~\citep{NLopt} (using Method of Moving Asymptotes~\citep{CCSA}) by solving a simplified version of~\eqref{eq:l0approx} with two priority levels, which effectively reduces to a NLP:
\begin{align}
	\min_{\chi,{\nu}_{2},{\tau}_{{2}}}\qquad &
	\textstyle\sum_{i\in\bC_2} \log(\tau_{i} + \xi)\\
	\text{s.t.}\qquad&
	-\tau_{i}\leq \nu_{i}\leq \tau_{i},\quad f_{i}(\chi) \geqq \nu_{i},\quad i\in\mathbb{C}_2\nonumber\\ 
	&f_{j}(\chi)\geq 0,\quad j\in\mI_1 \nonumber
\end{align}
 Unlike~\eqref{eq:l0approx}, the inequalities $f_{\mathcal{I}_{1}} \geq 0$ need to be feasible. 

UFactory's \texttt{xarm6} (end-effector $f_{\text{EF}}(\chi)$, $\chi\in\mathbb{R}^6$) needs to decide on which one of 10 to 100 objects to pick-up, while simultaneously computing the corresponding inverse kinematic posture (see Tab.~\ref{tab:xarm6shikp} and Fig.~\ref{fig:videosims} (a)). The inequality constraints $f_{j}$ represent joint angle limits and collision avoidance.

S-SHQP is shown in Fig.~\ref{fig:xarm6shikp} to reliably converge to a minimum decision-making error of less than $1\cdot 10^{-5}$~m. It can be observed that \ref{eq:shnlp} in combination with both the $\ell_1$-norm and $\ell_2$-norm leads to identically inaccurate results. This can be attributed to the quadratic formulation in~\eqref{eq:groupj}, which undermines the sparsity-promoting properties of the $\ell_1$-norm. This highlights the versatility and robustness of the $\ell_0$-norm, whose effectiveness is agnostic towards the specific problem formulation. Similarly, IPOPT manages to converge accurately for only 4 instances (10, 20, 40, 60 objects) at an approximate error of $3\cdot 10^{-8}$~m, while exceeding the number of maximum iterations of 3000 in the other cases. On the contrary, S-SHQP requires between 37 and 67 iterations for all samples. NLOPT finishes within 5 to 15 iterations, but manages to converge to a decision-making error of only $3$\,mm to $1.2$\,cm (solver tolerances: $1\cdot 10^{-5}$). All solvers use BFGS Lagrangian Hessian approximations.

Furthermore, the following representative MINLP is solved by SCIP~\citep{scip_minlp}:
\begin{align}
	 \min_{\chi,\theta_i} \qquad& \sum_{i=1}^{m}\theta_{i}\\
	 \text{s.t.}\qquad&
	 	\underline{\chi} \leq \chi\leq \overline{\chi}\nonumber\\
	 	& \Vert f_{\text{EF}}(\chi) - f_{\text{EF},d,i}\Vert_2^2 \leq 4\theta_i\nonumber\\
	 &\theta_i\in\{0,1\}\nonumber
	 \end{align}
	This example uses a planar robot with $n=2$ and link length $1$~m. The Big M scalar $4$ is derived from the reachable workspace of the robot. 
	Despite this simpler robot, the solver reaches time-out for 70 objects with $2.9\cdot 10^6$ branch-and-bound nodes explored, and 4854 IPOPT iterations for NLP relaxations (displayed in Fig.~\ref{fig:xarm6shikp}).

\subsection{Hierarchical decision-making with test functions}
\label{sec:eval:testfunc}
\begin{figure}[h!]
	\resizebox{1.\columnwidth}{!}   {%
		\begin{tabular}{@{} cccccccccccc @{}}  
			\toprule
			\vspace{-10pt}\\
			$\underline{l}$ & $\ell$ & $f_i(\chi)\leqq \nu_i,\quad i\in\bC_l$ & $\Vert (\nu_i^*) \Vert_2$ &  Iter.  \\
			\midrule
			$\underline{1}$  & 0  & $\chi_1^2 + \chi_2^2 - \nicefrac{1.9}{2} \leq \nu_{\nicefrac{i_1}{i_2}}$ & $\nicefrac{\color{Cerulean}\bm{0}}{\color{Cerulean}\bm{0}}$ & 1  \\
			\rowcolor{gray!20}
			$\underline{2}$ &  0  & $(1-\chi_1)^2 + 100(\chi_2 - \chi_1^2)^2 + \nicefrac{0}{5} = \nu_{\nicefrac{i_1}{i_2}}$ & $\nicefrac{2.9\cdot 10^{-4}}{5.0}$ & 72  \\
			$\underline{3}$ & 0  & $\chi_1^2 + \chi_2^2 - \nicefrac{0.9}{1} = \nu_{\nicefrac{i_1}{i_2}}$& $\nicefrac{1}{0.9}$ & 1  \\
			\rowcolor{gray!20}
			$\underline{4}$  & 0 & $\chi_2^2 + \chi_3^2 - \nicefrac{1}{1.1} = \nu_{\nicefrac{i_1}{i_2}}$& $\nicefrac{\color{Cerulean}\bm{0}}{0.1}$ & 7  \\
			$\underline{5}$ & 0  & $\chi_4^2 + \nicefrac{1}{1.1} \leq \nu_{\nicefrac{i_1}{i_2}}$& $\nicefrac{1}{ 1.1}$ & 7  \\
			\rowcolor{gray!20}
			$\underline{6}$ & 0  & $\chi_5^2 + \nicefrac{+1}{-1} \leq \nu_{\nicefrac{i_1}{i_2}}$& $\nicefrac{1}{ \color{Cerulean}\bm{0}}$ & 5  \\
			$\underline{7}$ & 0  & $\chi_6^2 + \chi_7^2 + \chi_8^2 - \nicefrac{4}{5} = \nu_{\nicefrac{i_1}{i_2}}$ & $\nicefrac{1}{ \color{Cerulean}\bm{1.3\cdot 10^{-8}}}$ & 4  \\
			\rowcolor{gray!20}
			$\underline{8}$  & 0  & $(1-\chi_6)^2 + 100(\chi_7 - \chi_6^2)^2 + \nicefrac{0}{4} = \nu_{\nicefrac{i_1}{i_2}}$ & $\nicefrac{\color{Cerulean}\bm{2.3\cdot 10^{-7}}}{4}$ & 2 \\
			$\underline{9}$  & 0  & $\sin(\chi_{_9} \hspace{-2pt}+\hspace{-2pt} \chi_{10}) \hspace{-2pt}+\hspace{-2pt} (\chi_9 \hspace{-2pt}-\hspace{-2pt} \chi_{10})^2 \hspace{-2pt}-
			\hspace{-2pt} 1.5x\chi_9 \hspace{-2pt}+\hspace{-2pt} 2.5\chi_{10} \hspace{-2pt}+\hspace{-2pt}1 \hspace{-2pt}=\hspace{-2pt} \nu_{i_1}$ & 21.9 & 29 \\
			& 0  & $\chi_9^2 + \chi_{10}^2 - 2 = \nu_{i_2}$ & \color{Cerulean}$\bm{2.8\cdot 10^{-17}}$ & \\
			\rowcolor{gray!20}
			$\underline{10}$  & 2 & $\chi = (\nu_{i})$& 3.01 & 1 \\
			\midrule
			$\Sigma$ & & & & 129  \\
			\bottomrule
		\end{tabular}
	}
	\captionof{table}{\ref{eq:shnlp} with $p=10$ and $n=10$, composed of test function selection constraints (Rosenbrock, etc.) as equalities and inequalities, solved by S-SHQP with $\mathcal{N}$\hspace{-2pt}QP. The norm $\ell=0,2$, optimal slacks $\nu^*$, and iteration counts (Iter.) are shown. On each priority level $l$, there are two constraints ${i_1}$ and ${i_2}$ summarized as ${\nicefrac{i_1}{i_2}}$.  Zero constraints are highlighted in blue.}
	
	\label{tab:p10nl}
\end{figure}

To validate solver behavior for hierarchical decision-making, we define a hierarchy of common optimization test functions (Rosenbrock, etc.) with $p=10$ levels and $n=10$ variables (Tab.~\ref{tab:p10nl}). The levels include identical (levels 1-8) or mixed (level 9) selection constraints. 
The results for S-SHQP with $\mathcal{N}$\hspace{-2pt}QP confirm the expected behavior:
\begin{itemize}
	\item Zero constraints are correctly identified for both equality (levels 2, 4, 7, 8, 9) and inequality (level 6) constraints.
	\item Infeasible selection constraints are removed to prevent Hessian activation (e.g., level 7), improving hierarchical decision-making on lower levels (see Sec.~\ref{sec:newtonm}).
\end{itemize}
%Due to the limited number of sparse constraints, \mbox{$\mathcal{N}$\hspace{-2pt}QP} ($130$~S-SHQP iterations / $36$~ms) and H-PIQP ($107$~S-SHQP iterations / $30$~ms) achieve comparable runtimes. H-MOSEK converges slower ($144$~S-SHQP iterations / $0.62$~s). 

\subsection{SHIK-P of humanoid robot with many candidate locations}
\label{sec:eval:plan}

\begin{table}[htp!]
	\centering
	\resizebox{1\columnwidth}{!}	{%
		\begin{tabular}{@{} cccccc @{}}  
			\toprule
			$\underline{l}$ & & $\ell$ & $f_i(\chi)\leqq \nu_i,\quad i\in\bC_l$ & $\Vert (\nu_i^*) \Vert_2$ &  Iter. \\
			\midrule
			$\underline{1}$ & Joint ang. lim.& 2 & {$\underline{\chi} \leq \chi \leq \overline{\chi}$} & 0  & 1  \\
			\rowcolor{gray!20}
			$\underline{2}$ & Coll. av. & 2 & $ f_{\text{coll}}(\chi) \geq (\nu_{\text{coll}})$ & $2.4\cdot 10^{-4}$ & 1 \\
			$\underline{3}$ & CoM$_z$ & 2 & $ f_{\text{CoM}_z}(\chi) - f_{\text{LF/RF},z}(\chi) - 0.2 \geq (\nu_{\text{CoM}})$ & $0$ & 1 \\
			\rowcolor{gray!20}
			\underline{4} & {LF } & 0 & $\Vert f_{\text{LF}}(\chi) - f_{\text{LF},d,s_1}(t)\Vert_2^2 = \nu_{s_1},\quad \vert\mathbb{S}_1\vert=200$ & ${3.8\cdot 10^{-6}}$ & 37 \\
			\rowcolor{gray!20}
			& {RF } & 0 & $\Vert f_{\text{RF}}(\chi) - f_{\text{RF},d,s_2}(t)\Vert_2^2 = \nu_{s_2},\quad \vert\mathbb{S}_2\vert=200$ & ${4.1\cdot 10^{-6}}$ &   \\
			$\underline{5}$ & LH  & 0 & $\Vert f_{\text{LH}}(\chi) - f_{\text{LH},d,s_1}(t)\Vert_2^2 = \nu_{s_1},\quad \vert\mathbb{S}_1\vert=200$ & $5.1\cdot 10^{-4}$ & 31 \\
			& {RH} & 0 & $\Vert f_{\text{RH}}(\chi) - f_{\text{RH},d,s_2}(t)\Vert_2^2 = \nu_{s_2},\quad \vert\mathbb{S}_2\vert=200$ &${6.5\cdot 10^{-5}}$ & \\
			\midrule
			$\Sigma$ & & & & & 71 \\
			\bottomrule
		\end{tabular}
	}
	\caption{\mbox{SHIK-P} of humanoid robot, $p=5$ and $n=31$, solved in $71$ iterations ($0.17$\,s, $\mathcal{N}$\hspace{-2pt}QP).}
	\label{tab:g1}
\end{table}
\begin{figure}[htp!]
	\includegraphics[width=1 \columnwidth]{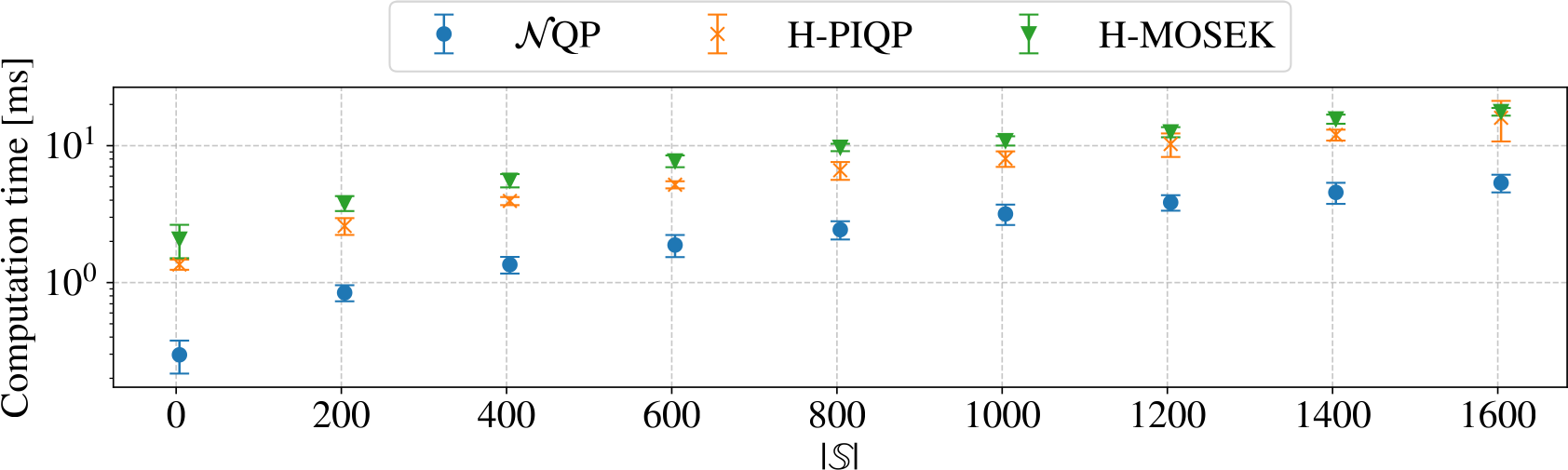}
	\centering
	\caption{Computation times of solving SHQP’s for humanoid robot inverse kinematics hierarchy (Tab.~\ref{tab:g1}, levels 4-5 only). The number of sparse equality constraints $\vert\mathbb{S}\vert$ is increased from $1$ to $1604$.}
	\label{fig:g1times}
\end{figure}

We compute the posture $\chi\in\mathbb{R}^{31}$ of the humanoid robot \texttt{G1} while autonomously selecting among Cartesian locations (see Fig.~\ref{fig:videosims} (b)). Each end-effector (LF, RF, LH, RH) is assigned 200 potential locations (Tab.~\ref{tab:g1}), for example identified by exteroceptive sensors. Unlike $\ell_2$-norm minimization, $\ell_0$ optimization enables choosing a single feasible location rather than averaging over all. The feet placement is thereby prioritized over the hand placement for hierarchical decision-making.

The hierarchy in Tab.~\ref{tab:g1} includes joint limits ($\underline{\chi}$, $\overline{\chi}$), collision avoidance (self and a horizontal bar), a center-of-mass (CoM) to feet minimum distance constraint in vertical direction to enforce balanced upright postures, and finally the selection constraints for each limb. The problem is solved in $0.23$~s ($89$~S-SHQP iterations) by \mbox{$\mathcal{N}$\hspace{-2pt}QP}.  
For comparison, H-PIQP converges in $90$~iterations / $0.33$~s, and H-MOSEK in $46$~iterations / $0.59$~s. Each end-effector successfully selects a unique feasible location, with results over $10$ different initial configurations visualized in Fig.~\ref{fig:videosims} (b). 

\begin{figure}[h!]
	\includegraphics[width=0.6\columnwidth]{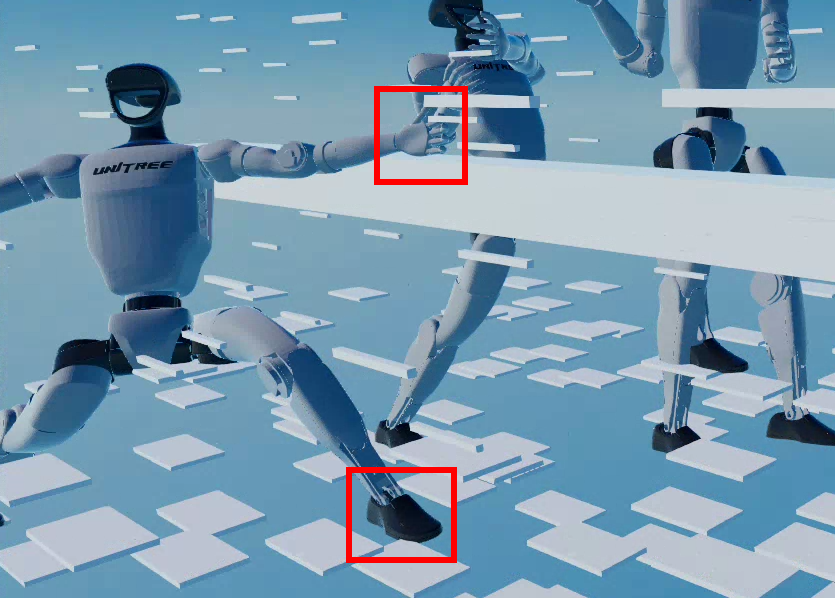}
	\centering
	\caption{S-SHQP is a local solver which can misguide the robot to reach for out-of-reach targets. Global understanding of the scene would encourage the robot to aim for feasible targets closer to the robot torso. Note that the rightmost robot posture has not converged yet.}
	\label{fig:shikpg1_infeas}
\end{figure}

As the proposed planner is local, some of the selected locations end up being infeasible due to conflict with same or higher priority constraints (see Fig.~\ref{fig:shikpg1_infeas}). 

To evaluate the impact of the selection constraint size $\vert\mathbb{S}\vert = 4,\dots,1604$ (corresponding to $1, \dots, 400$ possible locations per end-effector) on computation time, the hierarchy in Tab.~\ref{tab:g1} is solved considering only the equality constraints on levels~4 and~5. 
Figure~\ref{fig:g1times} confirms a linear relationship between computation time and the number of sparse $\ell_0$-norm constraints. 
The proposed $\mathcal{N}$\hspace{-2pt}QP solver consistently outperforms H-PIQP and H-MOSEK, both of which do not exploit the structured sparsity of SHQP problems.
\begin{figure}[tp!]
	\includegraphics[width=\columnwidth]{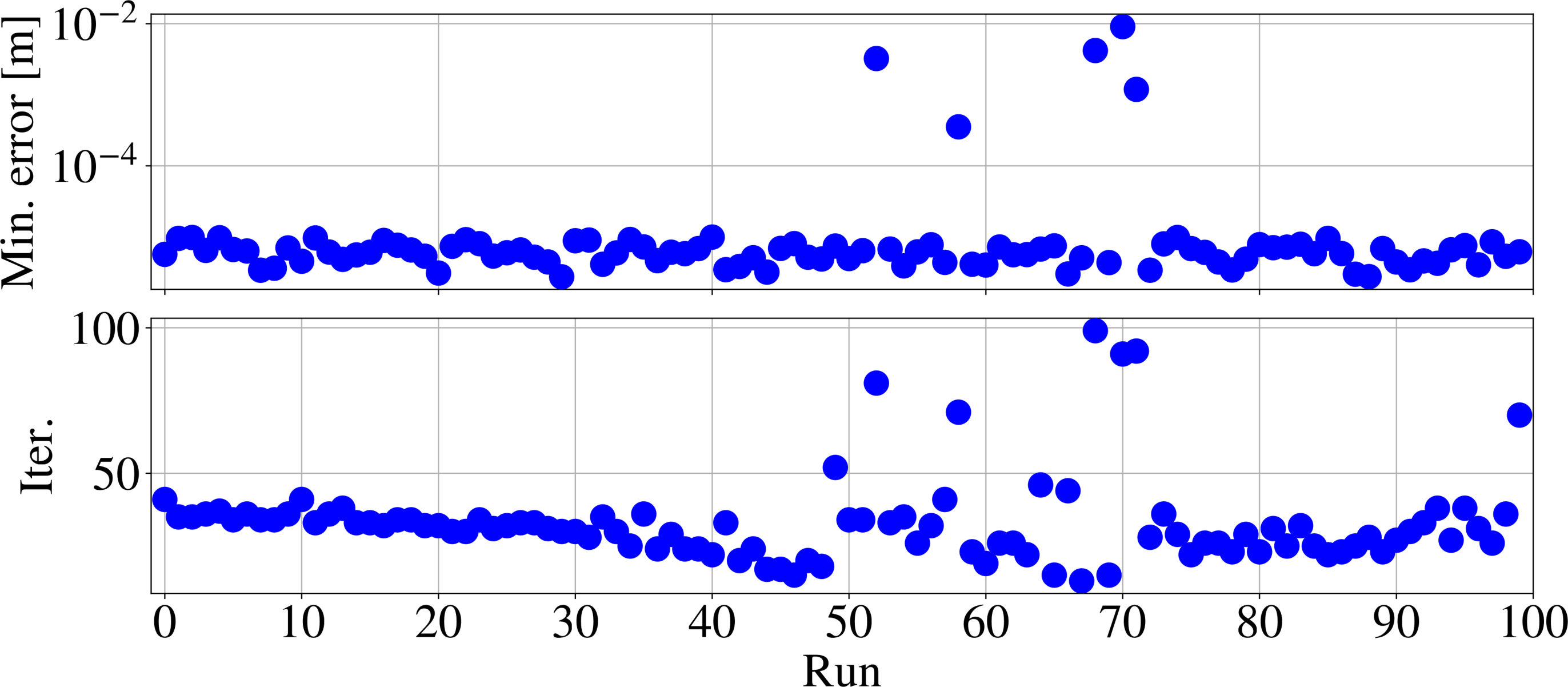}
	\centering
	\caption{S-SHQP is run a $100$\,times on level 4 only of hierarchy Tab.~\ref{tab:g1}.}
	\label{fig:shikp_g1_100runs}
\end{figure}

Lastly, we conduct 100 simulations on G1 foot location planning with different initial configurations in order to test the robustness of our solver S-SHQP. Infeasibility due to conflict with higher priority levels is avoided by solely considering level 4 of the hierarchy in Tab.~\ref{tab:g1}. The robot succeeds in 95 simulations with a DM error smaller than $1\cdot 10^{-5}$\,m on both feet (see Fig.~\ref{fig:shikp_g1_100runs}), while considering cost encoding via the mask in~\eqref{eq:l0mask} to avoid feet allocation to the same location. Similarly to 	Fig.~\ref{fig:shikpg1_infeas}, the failure cases can be explained by the local nature of the solver.
%We are not aware of any other method that addresses such parallel decision-making from the same set of candidates while avoiding redundant selection. 
%The mask~\eqref{eq:l0mask} could also be designed such that certain subsets of the decision space  are treated preferentially, for example when the robot should .

\subsection{\mbox{SHIK-C} of manipulator for object tracking}
\label{sec:eval:realtime}
\begin{table}[htp!]
	\centering
	\resizebox{0.9\columnwidth}{!}	{%
		\begin{tabular}{@{} cccccccc @{}}  
			\toprule
			$\underline{l}$ & & $\ell$ & $f_i(\chi)\leqq \nu_i,\quad i\in\bC_l$  \\
			\midrule
			$\underline{1}$ & Joint ang. lim. & 2 & $\underline{\chi} \leq \chi \leq \overline{\chi}$ \\
			\rowcolor{gray!20}
			$\underline{2}$ & Coll. av. & 2 & $ f_{\text{coll}}(\chi) \geq (\nu_{\text{coll}})$ 	\\
			$\underline{3}$ & CoM$_z$ & 2 & $ f_{\text{CoM}_z}(\chi) - f_{\text{Shoulder},z}(\chi) - 0.1 \geq (\nu_{\text{CoM}})$  \\
			\rowcolor{gray!20}
			$\underline{4}$ & EF & 0 & $\Vert f_{\text{EF}}(\chi) - f_{\text{EF},d,s_1}(t)\Vert_2^2 = \nu_{s_1},\quad\vert \mathbb{S}_1\vert=2$  \\
			\rowcolor{gray!20}
			& \color{blue}{Reg} & \color{blue}{0} & \color{blue}{${1\cdot 10^{-3}\chi = (\nu_{s_2})},\quad\vert \mathbb{S}_2\vert=n$} \\
			$\underline{5}$ & Reg. & 2 & $\chi =  \nu_{\text{Reg}}$ \\
			\bottomrule
		\end{tabular}
	}
	\caption{SHIK-C of manipulator: Task hierarchy for tracking of two targets with $p=5$ and $n=6$.}
	\label{tab:xarm6shikc}
\end{table}
\begin{figure}[htp!]
	\includegraphics[width=1\columnwidth]{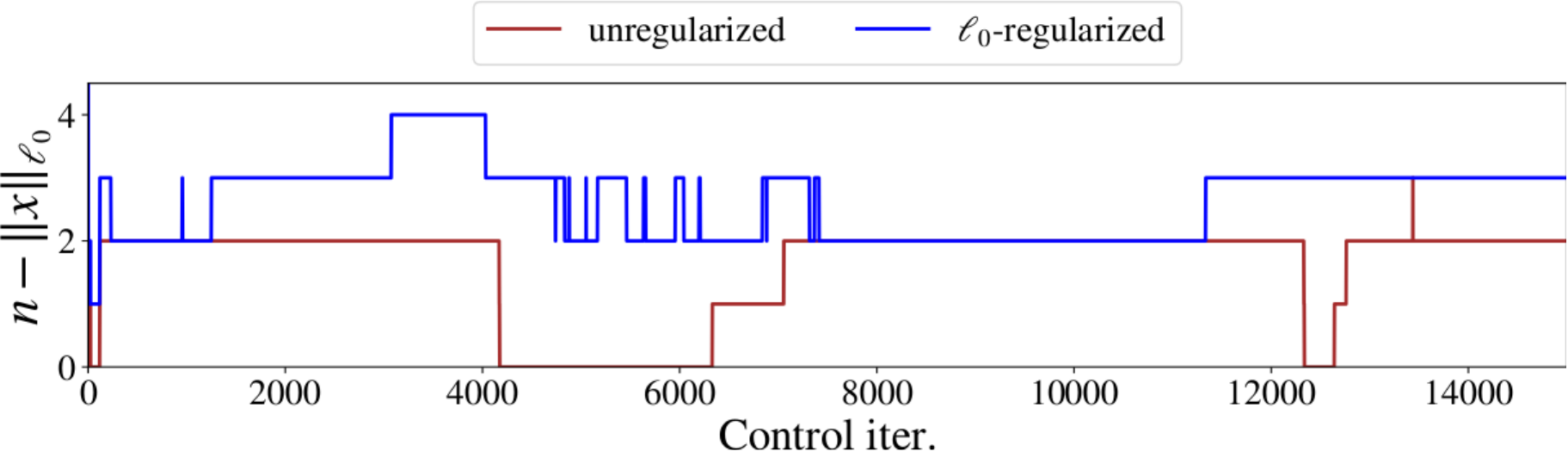}
	\centering
	\caption{SHIK-C of manipulator: Number of joints with zero angle for parsimonious kinematic control~\citep{Goncalves2015}.}
	\label{fig:xarm6eq}
\end{figure}

This test considers selective tracking of one of two moving targets with a real-world UFactory \texttt{xarm6} manipulator (see video). The two Cartesian locations $f_{\text{EF},d,i}(t)\in\mathbb{R}^3$ ($i=1,2$) move along opposing circular trajectories, partially occluded by a wall.  
The CoM of the robot must remain at least $0.1$~m above the shoulder link.  Optionally, an $\ell_0$-norm regularization term with small weight $1\cdot10^{-3}$ is added to achieve sparse kinematic control (colored in blue in Tab.~\ref{tab:xarm6shikc}). Such concurrent decision-making, parsimonious control, and presence of constraints is not possible in approaches like~\citep{Goncalves2015}.  

The control loop time is around $0.8$~ms using $\mathcal{N}$\hspace{-2pt}QP. Figure~\ref{fig:xarm6eq} shows the number of joints remaining at zero throughout execution. The $\ell_0$-regularized formulation leads to sparser actuation, e.g., joints $\chi_4$, $\chi_5$, and $\chi_6$ remain inactive when possible. Tracking accuracy decreases slightly, highlighting the challenge of adequately balancing between sparsity triggered by the logarithmic function in~\eqref{eq:l0approx}, and precision.
% Notably, placing the $\ell_0$-norm constraint at a lower hierarchy level is ineffective, as these tasks typically require full-rank Hessian regularization under Newton’s method for convergence.

\subsection{SHIK-C of humanoid robot with many candidate locations}
\label{sec:eval:continSelect}

\begin{table}[h!]
	\centering
	\resizebox{1\columnwidth}{!}	{%
		\begin{tabular}{@{} ccccc @{}}  
			\toprule
			$l$ & & $\ell$ & $f_i(\chi)\leqq \nu_i,\quad i\in\bC_l$  \\
			\midrule
			$\underline{1}$ & Joint ang. lim. & 2 & $\underline{\chi} \leq \chi \leq \overline{\chi}$   \\
			\rowcolor{gray!20}
			$\underline{2}$ & Coll. av. & 2 & $f_{\text{coll}}(\chi) \geq (\nu_{\text{coll}})$   \\
			$\underline{3}$ & LF / RF / LH & 2 & $ f_{i}(\chi) - f_{i,d}(t) = \nu_{i},\quad i\in \text{LF / RF / LH}$  \\
			\rowcolor{gray!20}
			$\underline{4}$ & RH & 0 / 1 / 2& $\Vert f_{\text{RH}}(\chi) - f_{\text{RH},d,s}(t)\Vert_2^2 = \nu_{s}$, $\vert\mathbb{S}\vert = 100$ \\
			$\underline{5}$ & Reg. & 2 & $\chi - \chi_d = \nu_{\text{Reg}}$   \\
			\bottomrule
		\end{tabular}
	}
	\caption{\mbox{SHIK-C} of humanoid robot with $p=5$ and $n=31$.}
	\label{tab:hrp2gsc}
\end{table} 

To assess solver performance in large-scale sparse control, we simulate a \texttt{G1} tasked with touching 100 objects moving downward, representing a dynamic catching scenario. The objects descend at constant velocity and are randomly distributed within reach of the robot.

The control hierarchy is shown in Tab.~\ref{tab:hrp2gsc}. End-effector location selection for the right-hand is formulated as a selection constraint with $\vert \mathbb{S}\vert = 100$ on level~4. When the distance to an object drops below $1$~cm, the object is removed from the group $\mathbb{S}$ for a continuous autonomous selection of new targets.

If the fourth level is formulated in the $\ell_0$-norm, the robot successfully interacts with 92 out of 100 objects before they pass beyond reach. If the fourth level is formulated as a linear or least-squares program ($\ell_1$ / $\ell_2$-norm), the robot fails to contact any (see Sec.~\ref{sec:sparseProgDecMak}). The \mbox{$\mathcal{N}$\hspace{-2pt}QP} solver handles each SHQP in $1.6\pm0.3$~ms, outperforming H-PIQP ($2.2\pm0.7$~ms) and H-MOSEK ($8.3\pm2.7$~ms). 
These results confirm the efficiency and scalability of the proposed sparse hierarchical formulation for both planning and real-time control tasks.

\subsection{Potential applications} %Robotic use-cases}
\label{sec:robotics}
%In e-commerce, sorting and distribution logistics or in...
%Two further simulations are given in the accompanying video where in one several manipulators are operating in a conveyor belt picking scenario while in the second a humanoid robot needs to place its hands on kinematically appropriate sides of randomly rotated boxes (see Fig.~\ref{fig:videosims}).

Sorting and distribution logistics play a critical role in maintaining competitiveness in e-commerce and other domains such as automated waste sorting. The following demonstrates how such tasks can be formulated as sparse optimization problems with immediate and continuous decision-making (SHIK-C), rather than relying on heuristic approaches which consider IK separately.

In the first simulation, multiple manipulators jointly address a shared selection constraint in which as many passing objects as possible must be removed from a conveyor (see Fig.~\ref{fig:videosims} (c)). Once an object has been removed by one of the robots, it is excluded from the selection constraint for all others. Such scenarios are representative of industrial sorting applications (e.g., e-commerce fulfillment or coal mine operations~\citep{convbeltgrasp25}), where different types of grasping end-effectors (such as suction cups, bi-manual grippers, or vacuum-based systems) may be employed depending on the nature of the objects.
\begin{table}[htp!]
	\centering
	\resizebox{0.9\columnwidth}{!}	{%
		\begin{tabular}{@{} cccccccc @{}}  
			\toprule
			$\underline{l}$ & & $\ell$ & $f_i(\chi)\leqq \nu_i,\quad i\in\bC_l$  \\
			\midrule
			$\underline{1}$ & Joint ang. lim. & 2 & $\underline{\chi} \leq \chi \leq \overline{\chi}$ \\
			\rowcolor{gray!20}
			$\underline{2}$ & Coll. av. & 2 & $ f_{\text{coll}}(\chi) \geq (\nu_{\text{coll}})$ 	\\
$\underline{4}$ & LF / RF & 2 & $\Vert f_{i}(\chi) - f_{i,d}\Vert_2^2 = \nu_{i},\quad i\in \text{LF / RF}$  \\
		\rowcolor{gray!20}
			$\underline{3}$ & CoM & 2 & $ f_{\text{CoM}}(\chi) \geq (\nu_{\text{CoM}})$  \\
			$\underline{4}$ & LH / RH & 0 & 
			$\left\Vert \BIN f_{\text{LH}}(\chi) - f_{\text{LH},d,s}(t)\Vert_2^2 
			\\f_{\text{RH}}(\chi) - f_{\text{LH}}(\chi)-f_{\text{RH},d}\BOUT\right\Vert_2^2= \nu_{s} ,\quad\vert \mathbb{S}\vert=4$  \\
			\rowcolor{gray!20}
			$\underline{5}$ & Reg. & 2 & $\chi =  \nu_{\text{Reg}}$ \\
			\bottomrule
		\end{tabular}
	}
	\caption{SHIK-C of humanoid robot: Task hierarchy for grasping randomly rotated box.}
	\label{tab:g1boxloading}
\end{table}

The second simulation illustrates how the two distinct left and right arm motions of a humanoid can be formulated within the same $\ell_0$-norm constraint (see Tab.~\ref{tab:g1boxloading}). Each of the four vertical sides of a box, placed on a table and randomly rotated within the range $[0, 2\pi]$, is assigned a unique identifier. The left hand is tasked with selecting one of the sides, while the right hand must be placed on the opposite side, Fig.~\ref{fig:videosims}~(d).

\section{Conclusion}

We have presented a sparse hierarchical optimization framework for hierarchical decision-making in model-based robotics. The~\ref{eq:shnlp} formulation integrates sparsity-promoting principles within hierarchical nonlinear optimization, enabling efficient planning and control of redundant robots under large sets of constraints and potential discrete candidate locations. 
By exploiting the structure of the quadratic sub-problem~\ref{eq:shqp}, the proposed numerical scheme achieves linear computational scaling with respect to the number of sparse constraints. The approach was validated on hierarchies composed of standard nonlinear test functions and demonstrated on robot \mbox{SHIK-P} and \mbox{SHIK-C} tasks involving autonomous location selection, confirming both convergence and computational efficiency.

This work opens a promising direction towards unifying model-based, whole-body optimal control and autonomous discrete contact planning by leveraging sparse hierarchical nonlinear optimization. We envision an incorporation into a reinforcement learning pipeline to enhance robustness against sparse gradients.
Notably, the $\ell_0$-norm approximation is differentiable, unlike other decision-making methods like mixed-integer programming, and is therefore amenable to informed training regimes like Sobolev training~\citep{cactosl}.

\section{Acknowledgements}

This work was partly supported by the RDF-24-02-060 XJTLU Research Development Fund.
This work was partly supported by the Schaeffler Hub for Advanced Research at Nanyang Technological University, under the ASTAR IAF-ICP Programme (grant number ICP1900093).

\bibliographystyle{plainnat}
\bibliography{bib}

\end{document}

%% file: commands.tex
\definecolor{light-gray}{gray}{0.95}
\definecolor{dark-gray}{gray}{0.5}
\definecolor{mygray}{gray}{0.75}
\newcommand{\BIN}{\begin{bmatrix}}
\newcommand{\BOUT}{\end{bmatrix}}

\newcommand{\bb}{\breve{b}}

\newcommand{\ta}{\tilde{a}}

\newcommand{\mA}{{\mathcal{A}}}

\newcommand{\mI}{{\mathcal{I}}}

\newcommand{\bC}{{\mathbb{C}}}
\newcommand{\bE}{{\mathbb{E}}}
\newcommand{\bI}{{\mathbb{I}}}

\newtheorem{prop}{Proposition}

\pgfdeclarelayer{bg1}   
\pgfdeclarelayer{bg}  
\pgfsetlayers{bg1,bg,main}  

\definecolor{orange}{rgb}{0.99,0.69,0.07}
\definecolor{lightgray}{gray}{0.85}
\definecolor{light-gray}{gray}{0.95}
\definecolor{dark-gray}{gray}{0.5}

\tikzset{cross/.style={cross out, draw=black, minimum size=2*(#1-\pgflinewidth), inner sep=0pt, outer sep=0pt},
cross/.default={1pt}}

 \makeatletter
 \newcommand\fs@spaceruled{\def\@fs@cfont{\bfseries}\let\@fs@capt\floatc@ruled
   \def\@fs@pre{\vspace{5pt}\hrule height.8pt depth0pt \kern2pt}%
   \def\@fs@post{\kern2pt\hrule\relax}%
   \def\@fs@mid{\kern2pt\hrule\kern2pt}%
   \let\@fs@iftopcapt\iftrue}
 \makeatother

\newtheorem{theorem}{Theorem}